\documentclass{article}

\usepackage{arxiv}
\usepackage[utf8]{inputenc} % allow utf-8 input
\usepackage[T1]{fontenc}    % use 8-bit T1 fonts
\usepackage{url}            % simple URL typesetting
\usepackage{booktabs}       % professional-quality tables
\usepackage{amsfonts,amsmath}       % blackboard math symbols
\usepackage{nicefrac}       % compact symbols for 1/2, etc.
\usepackage{microtype}      % microtypography
\usepackage{lipsum}
\usepackage{graphicx}
\graphicspath{ {./Figures/} }
\usepackage{subcaption}
\usepackage{array}
\usepackage{float}
\usepackage{cuted} 
\usepackage{amssymb}
\usepackage{mathtools,algorithm,algorithmicx,algpseudocode}

\usepackage[singlespacing]{setspace}
\usepackage{tocloft}

\usepackage[usenames,dvipsnames]{xcolor}
\definecolor{forceblue}{rgb}{0.2235, 0.4157, 0.6941}
\usepackage[colorlinks]{hyperref}
\AtBeginDocument{%
	\hypersetup{
		citecolor=forceblue,
		linkcolor=forceblue,   
		urlcolor=forceblue}}

\newcommand\blfootnote[1]{%
	\begingroup
	\renewcommand\thefootnote{}\footnote{#1}%
	\addtocounter{footnote}{-1}%
	\endgroup
}
\title{Stochastic Inference of Plate Bending from Heterogeneous Data: Physics-informed Gaussian Processes via Kirchhoff-Love theory}

\author{
	Igor Kavrakov$^{\star}$ \\
	University of Cambridge\\
	Cambridge, United Kingdom \\
	\And
	Gledson Rodrigo Tondo \\
	Bauhaus-Universität Weimar\\
	Weimar, Germany \\
	\And
	Guido Morgenthal \\
	Bauhaus-Universität Weimar\\
	Weimar, Germany \\
}

\begin{document}
%\tableofcontents
\maketitle
\begin{abstract}\blfootnote{Preprint published in \textit{ASCE J. Eng. Mech., 2025, 151(4): 04025005 (\href{https://doi.org/10.1061/JENMDT.EMENG-7558}{doi.org/10.1061/JENMDT.EMENG-7558})}}
Advancements in machine learning and an abundance of structural monitoring data have inspired the integration of mechanical models with probabilistic models to identify a structure's state and quantify the uncertainty of its physical parameters and response. In this paper, we propose an inference methodology for classical Kirchhoff-Love plates via physics-informed Gaussian Processes (GP). A probabilistic model is formulated as a multi-output GP by placing a GP prior on the deflection and deriving the covariance function using the linear differential operators of the plate governing equations. The posteriors of the flexural rigidity, hyperparameters, and plate response are inferred in a Bayesian manner using Markov chain Monte Carlo (MCMC) sampling from noisy measurements. We demonstrate the applicability with two examples: a simply supported plate subjected to a sinusoidal load and a fixed plate subjected to a uniform load. The results illustrate how the proposed methodology can be employed to perform stochastic inference for plate rigidity and physical quantities by integrating measurements from various sensor types and qualities. Potential applications of the presented methodology are in structural health monitoring and uncertainty quantification of plate-like structures.
\end{abstract}

%\begin{keyword}
%Gaussian Process Regression \sep Computational fluid dynamics \sep Vortex particle method \sep Buffeting \sep Long-span bridges
%\end{keyword}

\section{Introduction} \label{sec:Introduction}
%The classical plate model is one of the fundamental components of civil, mechanical or aerospace structures.
System identification of plate-like structures during health monitoring is a challenging task since there are inherent material and measurement uncertainties. Having noisy measurements, the goal is to determine the system's state by inferring uncertain physical parameters - plate stiffness or its potential change (damage) - and predict the system's response at locations without sensors~\cite{farrarIntroductionStructuralHealth2006a}.  Machine learning has been extensively applied in health monitoring~\cite{wordenApplicationMachineLearning2006}; however, often in a black-box manner without incorporating the mechanical model of the system. Another important aspect in identifying systems is fusing heterogeneous data from multiple sensor types/qualities in a physically meaningful manner~\cite{wuDataFusionApproaches2020} and quantifying the effect of their inherent uncertainties. This requires appropriate methodologies that leverage the system's mechanical laws to construct probabilistic machine learning models.\par

%Measurements obtained from such systems are heterogeneous, generally in the form of a structural response (deflections, rotations, strains), noisy and if obtained continuously, can generate a large amount of data. The goal of such models is usually to i) learn of the system parameters (e.g. stiffness) and quantify their uncertainty and ii) infer probabilistic physical quantities at unobserved locations or in time; both that can be later used to evaluate the system state. 

%The classical Kirchhoff-Love plates as mechanical models are across engineering field, such as  . Thus, they are used design and monitoring. Recent advances on sensor and computer technologies have vastly increased the amount of Structural Health Monitoring (SHM) systexms that are deployed in civil and mechanical structures. 

Recently, physics-informed machine learning models have been proposed for to integrate physical knowledge of the system and measurement data from its responses \cite{karniadakisPhysicsinformedMachineLearning2021,raissiHiddenPhysicsModels2018,raissiPhysicsinformedNeuralNetworks2019,fuhgIntervalFuzzyPhysicsinformed2022}. In contrast to the model updating methods (e.g. ~\cite{zhangTransferlearningGuidedBayesian2022,zhangStructuralDamageIdentification2021,monchettiBayesianbasedModelUpdating2022,sanayeiAutomatedFiniteElement2015}) that consider mechanical and data-driven model separately or purely data-driven techniques (e.g.~\cite{fangDamageIdentificationResponse2011,chakrabortyGraphtheoreticapproachassistedGaussianProcess2019,kavrakovDatadrivenAerodynamicAnalysis2022}), these models integrate a data-driven model (e.g. a neural network (NN) or Gaussian Process (GP) regression) within the governing partial differential equations (PDEs) of a mechanical model. The idea is to place the data-driven model on a specific physical quantity and leverage the governing PDEs directly on the data-driven model to arrive at the other physical quantities. This physics-informed model can then be trained based on available data and used for prediction. Commonly, NN are employed as regression models when dealing with physics-informed machine learning of PDEs, e.g. in fluid dynamics \cite{caiPhysicsinformedNeuralNetworks2021,guastoniConvolutionalnetworkModelsPredict2021,wangPhysicsinformedMachineLearning2017}, material science \cite{liuPhysicsinformedMachineLearning2021,weiMachineLearningPrediction2020,liuMultiFidelityPhysicsConstrainedNeural2019} and structural mechanics~\cite{vahabPhysicsInformedNeuralNetwork2022,yanFrameworkBasedPhysicsinformed2022,guoEnergyBasedErrorBound2022}.\par

In contrast to NN, GP regression~\cite{rasmussenGaussianProcessesMachine2006} is a non-parametric machine learning method with a powerful learning procedure that handles overfitting automatically. It is a Bayesian probabilistic model that infers distributions based on noisy observations by using flexible distribution of functions as prior through a covariance function. With certain restrictions, they can be viewed as infinitely deep neural networks \cite{nealPriorsInfiniteNetworks1996}. GPs have been used for input/output data-driven modeling in various applications, such as aerodynamics~\cite{kavrakovDatadrivenAeroelasticAnalyses2024} or spatiotemporal probabilistic reconstruction \cite{maProbabilisticReconstructionSpatiotemporal2022}. In a physics-informed setting, GPs were first used for solving stochastic differential equations by placing a GP prior on the forcing term and leveraging integro-differential operators to formulate the cross-covariance matrix in an analytical manner~\cite{sarkkaLinearOperatorsStochastic2011,alvarezLinearLatentForce2013}. The concept was extended by Raissi et al.~\cite{raissiMachineLearningLinear2017} who placed the GP prior on the solution of the differential equations, and included physical parameters as unknowns in the cross-covariance of the GP. This facilitates a powerful method capable of stochastic parameter inference and prediction based on noisy heterogeneous data. Physics-informed GPs and similar methods have been employed in system identification of Euler-Bernoulli and Timoshenko beams~\cite{tondoStochasticStiffnessIdentification2023,gregorySynthesisDataInstrumented2019}, dynamic force identification in mechanical systems~\cite{rogersLatentRestoringForce2022,tondoPhysicsinformedMachineLearning2023}, remote sensing~\cite{camps-vallsPhysicsawareGaussianProcesses2018} or modeling of ambient magnetic fields~\cite{solinModelingInterpolationAmbient2018}. \par 
In this study, we propose a physics-informed machine learning inference scheme for plates based on heterogeneous measurements, by combining the classical Kirchhoff-Love plate theory, physics-informed GPs and fully Bayesian inference using Markov chain Monte Carlo (MCMC) sampling~\cite{hastingsMonteCarloSampling1970,gelmanBayesianDataAnalysis2013}. A model is constructed as a multi-output GP of the plate physical quantities, i.e. deflections, strains, curvatures, load and internal forces. Specifically, we place the GP prior on deflections and derive analytical expressions for the remaining physical quantities that populate the covariance matrix. The flexural rigidity is also included in the covariance matrix as an unknown parameter. Having noisy heterogeneous measurement data, we train the model using the MCMC sampling and infer the probability distributions of the rigidity and kernel hyperparameters. These distributions can be then used to infer physical quantities at unobserved locations without any additional numerical discretisation or reguralisation techniques. Overall, the presented model provides uncertainty quantification of the parameters and prediction for heterogeneous sensors with varying quality.\par
The paper is organized as follows: First, the Kirchhoff-Love plate model is revisited to establish the physical laws governing the system. Next, the novel physics-informed GP model is presented, including learning and prediction strategies. Numerical experiments are then conducted for two examples of plates with different boundary conditions and loading, where the applicability of the method is demonstrated for various sensors setups and quality. Lastly, conclusions are made and the limitations are discussed.

\section{Kirchhoff-Love Plate Theory}\label{sec:2}
Classical theory for a thin plate with small deflections (see~Fig.~\ref{fig:Schematic_PlateTheory}) is described by a linear, fourth order PDE \cite{timoshenkoTheoryPlatesShells1987}:
\begin{equation}\label{eq:EqPlate}
	D\nabla^{4}w=D\left(\frac{\partial^{4}w}{\partial x^{4}}+\frac{2\partial^{4}w}{\partial x^{2}\partial y^{2}}+\frac{\partial^{4}w}{\partial y^{4}}\right)=q
\end{equation}
where $w=w(\boldsymbol{x})$ is the vertical (out-of-plane) deflection of a plate, which is described by $\boldsymbol{x}=(x,y)\in\mathbb{R}^2$. The load is denoted as $q=q(\boldsymbol{x})$ and $D$ is a constant that represents the flexural rigidity:
\begin{equation}
	D=\frac{Et^3}{12(1-\nu^2)},
\end{equation}
which is based on Young's modulus of elasticity $E$, plate thickness $t$ and Poisson ratio $\nu$. The vertical deflections are in linear differential relation with the rotations $r$ as:
\begin{equation}\label{eq:rotation}
	r_x=\frac{\partial w}{\partial x}, \, r_y=\frac{\partial w}{\partial y}, 
\end{equation}
and the curvatures $\kappa$ as
\begin{equation}\label{eq:Curvature}
	\kappa_x=-\frac{\partial^2 w}{\partial^2 x}, \, \kappa_y=-\frac{\partial^2 w}{\partial^2 y}, \, \kappa_{xy}=-2\frac{\partial^2 w}{\partial x\partial y}.
\end{equation}\par

The deflections, rotations, curvature and load are the typical measurable quantities when gathering data. From a design aspect, the interest lies in the internal forces, i.e. shear forces and bending moments. Based on the equilibrium equations, the moments $M$ in the corresponding direction are obtained as:
\begin{equation}\label{eq:Moments}
	\begin{aligned}
		M_x=-D\left(\frac{\partial^2w}{\partial x^2}+\nu\frac{\partial^2w}{\partial y^2}\right), \,
		M_y=-D\left(\frac{\partial^2w}{\partial y^2}+\nu\frac{\partial^2w}{\partial x^2}\right), \,
		M_{xy}= D(1-\nu)\frac{\partial^2 w}{\partial x \partial y},
	\end{aligned}
\end{equation} 
while the shear forces $Q$ as:
\begin{equation}\label{eq:Shear}
	\begin{aligned}
	Q_x=-D\frac{\partial}{\partial x}\nabla^2w=-D\frac{\partial}{\partial x}\left(\frac{\partial^2w}{\partial x^2}+\frac{\partial^2w}{\partial y^2}\right), \, Q_y=-D\frac{\partial}{\partial y}\nabla^2w=-D\frac{\partial}{\partial y}\left(\frac{\partial^2w}{\partial x^2}+\frac{\partial^2w}{\partial y^2}\right).
	\end{aligned}
\end{equation}\par 

Boundary conditions to the plate PDEs~\eqref{eq:EqPlate} can be applied by restricting the displacements and rotations at the edges depending on the support type
\begin{equation}\label{eq:BC_C}
	\begin{aligned}
	\text{Fixed edge:\hspace*{0.25cm}}	&w(\boldsymbol{x}_\mathrm{BC})=0,r_n(\boldsymbol{x}_\mathrm{BC})=0;\\
	\text{Simply-supported edge:\hspace*{0.25cm}} &w(\boldsymbol{x}_\mathrm{BC})=0.
	\end{aligned}
\end{equation}

\section{Physics-informed Gaussian Process Model of Kirchhoff-Love Plates} \label{sec:3}
\subsection{Problem statement}\label{sec:ProbStat}
Consider noisy measurements $\boldsymbol{z}=\left\{z_{i}\right\}_{i=1}^{N}\in\mathbb{R}^{N\times1}$ of the plate quantities at measurement locations $\boldsymbol{X}=\left\{\boldsymbol{x}_{i}\right\}_{i=1}^{N}\in\mathbb{R}^{N\times2}$, where a particular location is denoted as $\boldsymbol{x}_i=(x_i,y_i)$ for $i=1,\dots,N$ and $N$ being the number of measurement locations. Based on the heterogeneous measurement data set $\mathcal{D}=\left\{\boldsymbol{X},\boldsymbol{z}\right\}$ and the mechanical relations~\eqref{eq:EqPlate}-\eqref{eq:BC_C}, the objectives of the proposed model are two fold:
\begin{itemize}
	\item[i)] to infer the distribution of the flextural rigidity $D$;
	\item[ii)] to infer the predictive distributions of plate quantities at prediction points $\boldsymbol{X}^*$.
\end{itemize}\par

The measurements can be of any physical quantity (deflections, rotations, curvatures or loads) at disassociated points. For each quantity - example, deflection - we consolidate the $N_w$ number of measurement observations $z_{w,i}$ into the vector $\boldsymbol{z}_w\in\mathbb{R}^{N_w\times1}$ at coordinate points $\boldsymbol{x}_{w,i}$ assembled in $\boldsymbol{X}_w\in\mathbb{R}^{N\times2}$. In this way, different types of measurements and their corresponding points can be consolidated into vectors and matrices. Example, in case of the deflections and load, the measurement vector is $\boldsymbol{z}_{wq}=(\boldsymbol{z}_w,\boldsymbol{z}_q)\in\mathbb{R}^{N_{wq}\times1}$ and the corresponding measurement points  $\boldsymbol{X}_{wq}=(\boldsymbol{X}_w;\boldsymbol{X}_q)\in\mathbb{R}^{N_{wq}\times2}$ for $N_{wq}=N_w+N_q$. Similar notation is used for the other quantities and when all quantities are considered, the subscripts are dropped; e.g., $\boldsymbol{z}$ contains all physical quantities.\par
In what follows, we begin by formulating the physics-informed model. The leaning of both the flexural rigidity and statistical hyperparameters is described next (objective i), followed by formulation of the predictive posteriors (objective ii).
\subsection{Model formulation}\label{sec:ModelForm}

Considering at first the deflection $w$ and load $q$ over the field $\boldsymbol{x}$, we assume a zero-mean GP {\textit{prior}} on the deflections:
\begin{equation}\label{eq:GP_w}
	w(\boldsymbol{x})\sim\mathcal{GP}(0,k_{ww}(\boldsymbol{x},\boldsymbol{x}^\prime;\boldsymbol{\theta})),
\end{equation}
with zero mean and a covariance function $k_{ww}=k_{ww}(\boldsymbol{x},\boldsymbol{x}^\prime;\boldsymbol{\theta})$ that is a function of specific hyperparameters $\boldsymbol{\theta}$. The covariance function (i.e. kernel) encodes the prior statistical assumptions of the form of the latent function to be inferred, and will be discussed later. The deflections and load are in a linear relationship $\mathcal{L}_{qx}^{D}w(\boldsymbol{x})=q(\boldsymbol{x})$ according to~\eqref{eq:EqPlate}, where the operator $\mathcal{L}_{qx}^{D}=D\nabla^4$ also contains the flexural rigidity $D$ as a physical parameter. Since a linear transformation of a GP is also a GP~\cite{sarkkaLinearOperatorsStochastic2011}, the load $q$ can be obtained as
\begin{equation}\label{eq:GP_q}
	q(\boldsymbol{x})\sim\mathcal{GP}(0,k_{qq}^D(\boldsymbol{x},\boldsymbol{x}^\prime;\boldsymbol{\theta},D)),
\end{equation}
with a covariance function $k_{qq}$ related to $k_{ww}$ through the mechanical plate model, i.e. the operator $\mathcal{L}_{qx}$ as:
\begin{equation}\label{eq:Cov} k_{qq}^D(\boldsymbol{x},\boldsymbol{x}^\prime;\boldsymbol{\theta},D)=\mathcal{L}_{qx}^D\mathcal{L}_{qx^\prime}^Dk_{ww}(\boldsymbol{x},\boldsymbol{x}^\prime;\boldsymbol{\theta}).
\end{equation}
The load and deflections are heterogeneous output of the same prior, which yields a multi-output GP:
	\begin{equation}\label{eq:MultiOutput_GP}
		\begin{aligned}
			\boldsymbol{\varphi}_{wq}(\boldsymbol{x})=					\begin{bmatrix}
				w(\boldsymbol{x})\\
				q(\boldsymbol{x})\\
			\end{bmatrix}&\sim\mathcal{GP}\left(\boldsymbol{0},\boldsymbol{\mathcal{K}}_{wq}(\boldsymbol{x},\boldsymbol{x}^\prime;\boldsymbol{\theta},D)\right)\\
&=\mathcal{GP}\left(\boldsymbol{0},\begin{bmatrix}
			k_{ww}(\boldsymbol{x},\boldsymbol{x}^\prime;\boldsymbol{\theta}) & k_{wq}^D(\boldsymbol{x},\boldsymbol{x}^\prime;\boldsymbol{\theta},D)\\
			k_{qw}^D(\boldsymbol{x},\boldsymbol{x}^\prime;\boldsymbol{\theta},D) & k_{qq}^D(\boldsymbol{x},\boldsymbol{x}^\prime;\boldsymbol{\theta},D)\\
		\end{bmatrix}\right),
	\end{aligned}
	\end{equation}
 where the cross-covariance between ${w}(\boldsymbol{x})$ and ${q}(\boldsymbol{x}^\prime)$, and, correspondingly, between ${w}(\boldsymbol{x}^\prime)$ and ${q}(\boldsymbol{x})$, are obtained as
 \begin{equation}\label{eq:CrossCov}
	\begin{aligned}
	\displaystyle 
		 k_{qw}^D(\boldsymbol{x},\boldsymbol{x}^\prime;\boldsymbol{\theta},D)&=\mathcal{L}_{qx}^Dk_{ww}(\boldsymbol{x},\boldsymbol{x}^\prime;\boldsymbol{\theta}),\\	
		k_{wq}^D(\boldsymbol{x},\boldsymbol{x}^\prime;\boldsymbol{\theta},D)&=\mathcal{L}_{qx^\prime}^Dk_{ww}(\boldsymbol{x},\boldsymbol{x}^\prime;\boldsymbol{\theta}),
	\end{aligned}
\end{equation}
respectively. The cross-covariance functions impose non-trivial correlation between the load and deflections regressions based on physical laws. Here, it is important to note the notation $\mathcal{L}_{qx}^D$ means applying the operator to the covariance $k_{ww}$ w.r.t. first argument, $\boldsymbol{x}$, while the  notation $\mathcal{L}_{qx^\prime}^D$ means applying the operator w.r.t. second argument, $\boldsymbol{x}^\prime$. This convention is required to cope with anti-symmetric cross-covariance that may result from the operator and selected covariance function. The superscript $D$ denotes that the operator includes the flexural rigidity.\par
Using the same logic, the rotations, curvatures, and internal forces are also GPs. Thus, the complete physics-informed GP \textit{prior} based on classical plate theory is a multi-output GP: 
\begin{equation}\label{eq:MultiOutput_GPFull}
		\boldsymbol{\varphi}(\boldsymbol{x})\sim\mathcal{GP}\left(\boldsymbol{0},\boldsymbol{\mathcal{K}}(\boldsymbol{x},\boldsymbol{x}^\prime;\boldsymbol{\theta},D)\right),
\end{equation}
where $\boldsymbol{\varphi}=(w,r_x,r_y,\kappa_x,\kappa_y,\kappa_{xy},q,Q_x,Q_y,M_x,M_y,M_{xy})^T$ is a vector function. The matrix function $\boldsymbol{\mathcal{K}}$ is comprised of covariance functions based on $k_w$, and~\eqref{eq:Cov} and~\eqref{eq:CrossCov}, where the corresponding linear operators (cf.~\eqref{eq:rotation} to~\eqref{eq:Shear}) are employed: 
\begin{equation}\label{eq:Operators}
	\begin{array}{lll}
		\displaystyle
		\mathcal{L}_{r_{x}}=\frac{\partial}{\partial x}, &\displaystyle \mathcal{L}_{r_{y}}=\frac{\partial}{\partial y}, &\displaystyle
		\mathcal{L}_{\kappa_{x}}=-\frac{\partial^2}{\partial x^2},\\
		\displaystyle  \mathcal{L}_{\kappa_{y}}=-\frac{\partial^2}{\partial y^2},  &\displaystyle \mathcal{L}_{\kappa_{xy}}=-2\frac{\partial^2}{\partial x \partial y},&\displaystyle \mathcal{L}_{q}^D=D\nabla^4,\\
		\displaystyle  \mathcal{L}_{Q_{x}}^D=-D\frac{\partial}{\partial x}\nabla^2,  &\displaystyle  \mathcal{L}_{Q_{y}}^D=-D\frac{\partial}{\partial y}\nabla^2,&\displaystyle\mathcal{L}_{M_{x}}^D=-D\left(\frac{\partial^2}{\partial x^2}+\nu\frac{\partial^2}{\partial y^2}\right), \\
		\multicolumn{3}{c}{
			\begin{array}{ll}
				\displaystyle   \displaystyle  \mathcal{L}_{M_{y}}^D=-D\left(\frac{\partial^2}{\partial y^2}+\nu\frac{\partial^2}{\partial x^2}\right), & \displaystyle \mathcal{L}_{M_{xy}}^D=D(1-\nu)\frac{\partial^2}{\partial x \partial y}.
		\end{array}}
	\end{array}
\end{equation}
Example, the cross-covariance between the curvature $\kappa_x$ and load $q$ can be obtained as follows:
\begin{equation}\label{eq:CrossCov1}
	\begin{aligned}
		\displaystyle 	k_{\kappa_xq}^D(\boldsymbol{x},\boldsymbol{x}^\prime;\boldsymbol{\theta},D)&=\mathcal{L}_{\kappa_xx}\mathcal{L}_{qx^\prime}^Dk_{ww}(\boldsymbol{x},\boldsymbol{x}^\prime;\boldsymbol{\theta}),\\
		 k_{q\kappa_x}^D(\boldsymbol{x},\boldsymbol{x}^\prime;\boldsymbol{\theta},D)&=\mathcal{L}_{qx}^D\mathcal{L}_{\kappa_xx^\prime}k_{ww}(\boldsymbol{x},\boldsymbol{x}^\prime;\boldsymbol{\theta}).
	\end{aligned}
\end{equation} 
The rest of the cross-covariance functions can be obtained in a similar manner and explicit relations are derived in Appendix~\ref{App:CovFunc}.\par
The prior in~\eqref{eq:MultiOutput_GPFull} is defined over a continuous field $\boldsymbol{x}$, whereas we are usually interested in discrete points where measurements are available $\boldsymbol{X}$ or prediction is warranted $\boldsymbol{X}^*$. The \textit{prior} in a discrete sense is then a collection of random variables $\boldsymbol{f}=\left\{\boldsymbol{\varphi}(\boldsymbol{x}_i)\right\}_{i=1}^{N}\in\mathbb{R}^{N\times1}$ (e.g. deflection at discrete locations $\boldsymbol{f}_w=w(\boldsymbol{X}_w)$), which based on~\eqref{eq:MultiOutput_GPFull} results in a multivariate normal distribution:
\begin{equation}\label{eq:MultiOutput_GP_PlateShort}
	\boldsymbol{f}\sim p(\boldsymbol{f}|\boldsymbol{X},\boldsymbol{\theta},D)=\mathcal{N}(\boldsymbol{0},\boldsymbol{K}).
\end{equation}
Given in an expanded form for the discrete prior $\boldsymbol{f}$ and covariance matrix $\boldsymbol{K}$ this yields:

\AtBeginEnvironment{bmatrix}{\setlength{\arraycolsep}{1pt}}
%\begin{singlespace}
	\begin{equation}\label{eq:MultiOutput_GP_Plate}
		\hspace*{-1.0cm}\footnotesize\begin{bmatrix}
			\boldsymbol{f}_{w}\\
			\boldsymbol{f}_{r_{x}}\\
			\boldsymbol{f}_{r_{y}}\\
			\boldsymbol{f}_{\kappa_x}\\
			\boldsymbol{f}_{\kappa_y}\\
			\boldsymbol{f}_{\kappa_{xy}}\\
			\boldsymbol{f}_{q}\\
			\boldsymbol{f}_{Q_x}\\
			\boldsymbol{f}_{Q_y}\\
			\boldsymbol{f}_{M_x}\\
			\boldsymbol{f}_{M_y}\\
			\boldsymbol{f}_{M_{xy}}\\
		\end{bmatrix} \! \! \! \sim \! \mathcal{N} \! \left(\! \! \boldsymbol{0}, \! \! \begin{bmatrix}
			\boldsymbol{K}_{ww} & \boldsymbol{K}_{wr_x} & \boldsymbol{K}_{wr_y} & \boldsymbol{K}_{w\kappa_x} & \boldsymbol{K}_{w\kappa_y} & \boldsymbol{K}_{w\kappa_{xy}} & \boldsymbol{K}_{wq}^D& \boldsymbol{K}_{wQ_x}^D & \boldsymbol{K}_{wQ_y}^D & \boldsymbol{K}_{wM_x}^D & \boldsymbol{K}_{wM_y}^D & \boldsymbol{K}_{wM_{xy}}^D\\
			
			\boldsymbol{K}_{r_xw} & \boldsymbol{K}_{r_xr_x} & \boldsymbol{K}_{r_xr_y} & \boldsymbol{K}_{r_x\kappa_x} & \boldsymbol{K}_{r_x\kappa_y} & \boldsymbol{K}_{r_x\kappa_{xy}} & \boldsymbol{K}_{r_xq}^D & \boldsymbol{K}_{r_xQ_x}^D & \boldsymbol{K}_{r_xQ_y}^D & \boldsymbol{K}_{r_xM_x}^D & \boldsymbol{K}_{r_xM_y}^D & \boldsymbol{K}_{r_xM_{xy}}^D\\
			
			\boldsymbol{K}_{r_yw} & \boldsymbol{K}_{r_yr_x} & \boldsymbol{K}_{r_yr_y} & \boldsymbol{K}_{r_y\kappa_x} & \boldsymbol{K}_{r_y\kappa_y} & \boldsymbol{K}_{r_y\kappa_{xy}} & \boldsymbol{K}_{r_yq}^D& \boldsymbol{K}_{r_yQ_x}^D & \boldsymbol{K}_{r_yQ_y}^D & \boldsymbol{K}_{r_yM_x}^D & \boldsymbol{K}_{r_yM_y}^D & \boldsymbol{K}_{r_yM_{xy}}^D\\		
			
			\boldsymbol{K}_{\kappa_xw} & \boldsymbol{K}_{\kappa_xr_x} & \boldsymbol{K}_{\kappa_xr_y} & \boldsymbol{K}_{\kappa_x\kappa_x} & \boldsymbol{K}_{\kappa_x\kappa_y} & \boldsymbol{K}_{\kappa_x\kappa_{xy}} & \boldsymbol{K}_{\kappa_xq}^D& \boldsymbol{K}_{\kappa_xQ_x}^D & \boldsymbol{K}_{\kappa_xQ_y}^D & \boldsymbol{K}_{\kappa_xM_x}^D & \boldsymbol{K}_{\kappa_xM_y}^D & \boldsymbol{K}_{\kappa_xM_{xy}}^D\\
			
			\boldsymbol{K}_{\kappa_yw} & \boldsymbol{K}_{\kappa_yr_x} & \boldsymbol{K}_{\kappa_yr_y} & \boldsymbol{K}_{\kappa_y\kappa_x} & \boldsymbol{K}_{\kappa_y\kappa_y} & \boldsymbol{K}_{\kappa_y\kappa_{xy}} & \boldsymbol{K}_{\kappa_yq}^D& \boldsymbol{K}_{\kappa_yQ_x}^D & \boldsymbol{K}_{\kappa_yQ_y}^D & \boldsymbol{K}_{\kappa_yM_x}^D & \boldsymbol{K}_{\kappa_yM_y}^D & \boldsymbol{K}_{\kappa_yM_{xy}}^D\\	
			
			\boldsymbol{K}_{\kappa_{xy}w} & \boldsymbol{K}_{\kappa_{xy}r_x} & \boldsymbol{K}_{\kappa_{xy}r_y} & \boldsymbol{K}_{\kappa_{xy}\kappa_x} & \boldsymbol{K}_{\kappa_{xy}\kappa_y} & \boldsymbol{K}_{\kappa_{xy}\kappa_{xy}} & \boldsymbol{K}_{\kappa_{xy}q}^D& \boldsymbol{K}_{\kappa_{xy}Q_x}^D & \boldsymbol{K}_{\kappa_{xy}Q_y}^D & \boldsymbol{K}_{\kappa_{xy}M_x}^D & \boldsymbol{K}_{\kappa_{xy}M_y}^D & \boldsymbol{K}_{\kappa_{xy}M_{xy}}^D\\	
			
			\boldsymbol{K}_{qw}^D & \boldsymbol{K}_{qr_x}^D & \boldsymbol{K}_{qr_y}^D & \boldsymbol{K}_{q\kappa_x}^D & \boldsymbol{K}_{q\kappa_y}^D & \boldsymbol{K}_{q\kappa_{xy}}^D & \boldsymbol{K}_{qq}^D& \boldsymbol{K}_{qQ_x}^D & \boldsymbol{K}_{qQ_y}^D & \boldsymbol{K}_{qM_x}^D & \boldsymbol{K}_{qM_y}^D & \boldsymbol{K}_{qM_{xy}}^D\\
			
			\boldsymbol{K}_{Q_xw}^D & \boldsymbol{K}_{Q_xr_x}^D & \boldsymbol{K}_{Q_xr_y}^D & \boldsymbol{K}_{Q_x\kappa_x}^D & \boldsymbol{K}_{Q_x\kappa_y}^D & \boldsymbol{K}_{Q_x\kappa_{xy}}^D & \boldsymbol{K}_{Q_xq}^D& \boldsymbol{K}_{Q_xQ_x}^D & \boldsymbol{K}_{Q_xQ_y}^D & \boldsymbol{K}_{Q_xM_x}^D & \boldsymbol{K}_{Q_xM_y}^D & \boldsymbol{K}_{Q_xM_{xy}}^D\\
			
			\boldsymbol{K}_{Q_yw}^D & \boldsymbol{K}_{Q_yr_x}^D & \boldsymbol{K}_{Q_yr_y}^D & \boldsymbol{K}_{Q_y\kappa_x}^D & \boldsymbol{K}_{Q_y\kappa_y}^D & \boldsymbol{K}_{Q_y\kappa_{xy}}^D & \boldsymbol{K}_{Q_yq}^D& \boldsymbol{K}_{Q_yQ_x}^D & \boldsymbol{K}_{Q_yQ_y}^D & \boldsymbol{K}_{Q_yM_x}^D & \boldsymbol{K}_{Q_yM_y}^D & \boldsymbol{K}_{Q_yM_{xy}}^D\\
			
			\boldsymbol{K}_{M_xw}^D & \boldsymbol{K}_{M_xr_x}^D & \boldsymbol{K}_{M_xr_y}^D & \boldsymbol{K}_{M_x\kappa_x}^D & \boldsymbol{K}_{M_x\kappa_y}^D & \boldsymbol{K}_{M_x\kappa_{xy}}^D & \boldsymbol{K}_{M_xq}^D& \boldsymbol{K}_{M_xQ_x}^D & \boldsymbol{K}_{M_xQ_y}^D & \boldsymbol{K}_{M_xM_x}^D & \boldsymbol{K}_{M_xM_y}^D & \boldsymbol{K}_{M_xM_{xy}}^D\\
			
			\boldsymbol{K}_{M_yw}^D & \boldsymbol{K}_{M_yr_x}^D & \boldsymbol{K}_{M_yr_y}^D & \boldsymbol{K}_{M_y\kappa_x}^D & \boldsymbol{K}_{M_y\kappa_y}^D & \boldsymbol{K}_{M_y\kappa_{xy}}^D & \boldsymbol{K}_{M_yq}^D& \boldsymbol{K}_{M_yQ_x}^D & \boldsymbol{K}_{M_yQ_y}^D & \boldsymbol{K}_{M_yM_x}^D & \boldsymbol{K}_{M_yM_y}^D & \boldsymbol{K}_{M_yM_{xy}}^D\\
			
			\boldsymbol{K}_{M_{xy}w}^D & \boldsymbol{K}_{M_{xy}r_x}^D & \boldsymbol{K}_{M_{xy}r_y}^D & \boldsymbol{K}_{M_{xy}\kappa_x}^D & \boldsymbol{K}_{M_{xy}\kappa_y}^D & \boldsymbol{K}_{M_{xy}\kappa_{xy}}^D & \boldsymbol{K}_{M_{xy}q}^D& \boldsymbol{K}_{M_{xy}Q_x}^D & \boldsymbol{K}_{M_{xy}Q_y}^D & \boldsymbol{K}_{M_{xy}M_x}^D & \boldsymbol{K}_{M_{xy}M_y}^D & \boldsymbol{K}_{M_{xy}M_{xy}}^D\\		
		\end{bmatrix} \! \right).
	\end{equation}
\begin{figure}[!b]
	\centering
	\includegraphics[clip,width=\columnwidth]{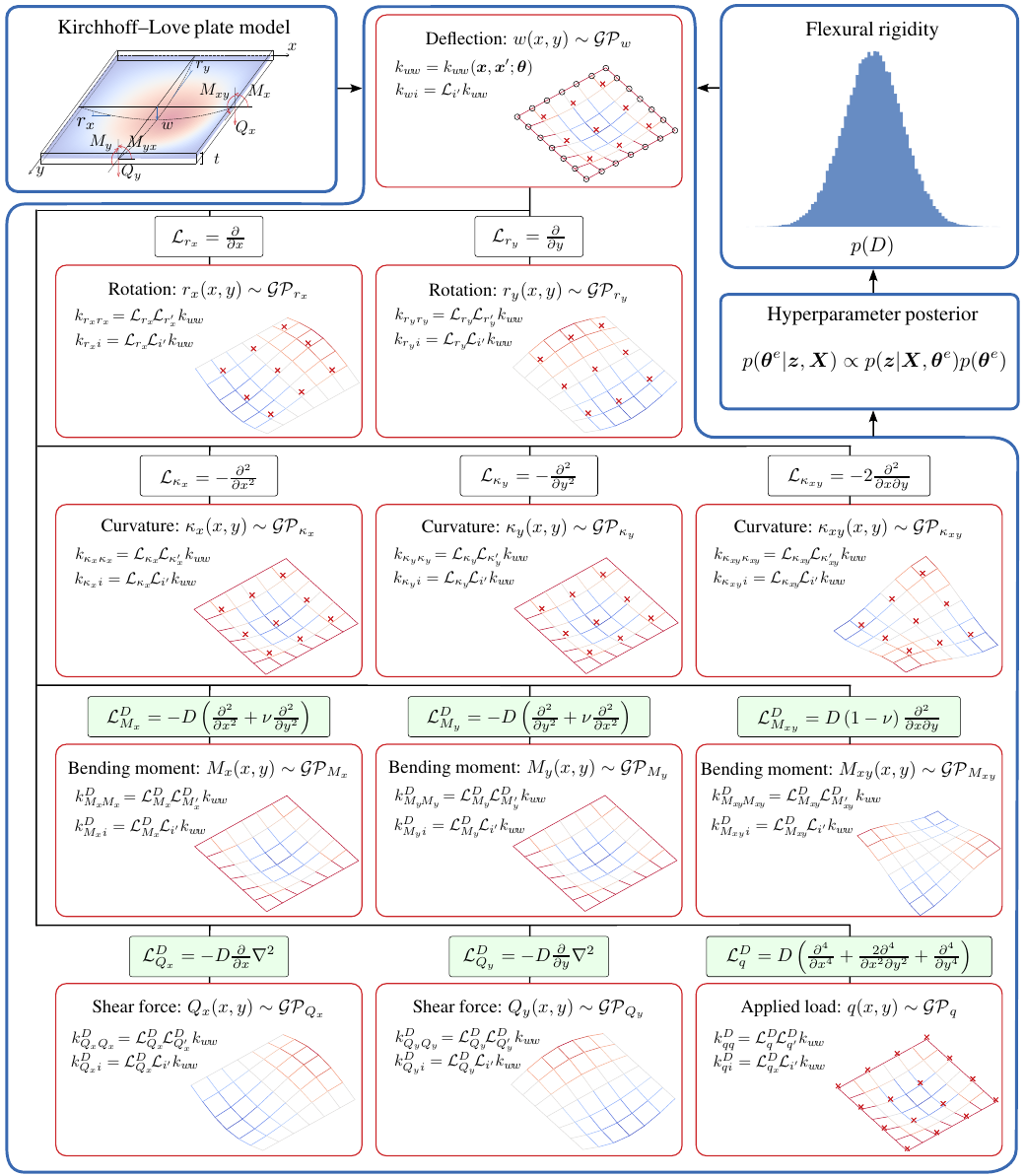} 
	\caption{Schematic for a physics-informed GP model: The model is derived leveraging the Kirchhoff-Love theory for the assumed GP prior on the deflection $w$. The flexural rigidity $D$ (green operators) is part of the model formulation and it can be inferred from noisy observations $\boldsymbol{z}$ (red crosses). The boundary conditions are imposed as noiseless observations $\boldsymbol{z}_\mathrm{BC}$ (black crosses).}
	\label{fig:Schematic_PlateTheory}
\end{figure} 
The covariance matrix $\boldsymbol{K}=\boldsymbol{\mathcal{K}}(\boldsymbol{X},\boldsymbol{X};\boldsymbol{\theta},D)$ is now a block matrix, which members can be obtained based on $k_{ww}$ and the physics-informed covariance functions defined in Appendix~\ref{App:CovFunc} at the corresponding locations. Example, the block entries for the load and deflection are (dropping the dependence on $\boldsymbol{\theta}$ and $D$ in notation):  $\boldsymbol{K}_{ww}=k_{ww}(\boldsymbol{X}_w,\boldsymbol{X}_w)$,  $\boldsymbol{K}_{wq}^D=k_{wq}^D(\boldsymbol{X}_w,\boldsymbol{X}_q)$, $\boldsymbol{K}_{qw}^D=k_{qw}^D(\boldsymbol{X}_q,\boldsymbol{X}_w)$ and   $\boldsymbol{K}_{qq}^D=k_{qq}^D(\boldsymbol{X}_q,\boldsymbol{X}_q)$. \par 
The covariance matrix $\boldsymbol{K}$ is constructed based on the selected deflection covariance function $k_{ww}$. This is the only term in the matrix $\boldsymbol{K}$ that is predefined, while the others are determined by the plate governing equation. Selecting $k_{ww}$ encodes the statistical assumptions or knowledge (e.g., stationarity, smoothness) in the prior, in addition to the mechanics introduced by the operators $\mathcal{L}$ for the other covariance functions. Here, we use the squared exponential covariance function with Automatic Relevance Determination (ARD)~\cite{nealPriorsInfiniteNetworks1996,mackay1998introduction}.  Using the exponential function assumes a continuous deflection field that is infinitely differentiable, allowing continuity up to the continuous load $q$, for which the operator has the highest order of differentiation (fourth order). The ARD property offers anisotropy, which is particularly important for physical quantities that are not reflection-symmetric with respect to the axes (e.g., curvatures).\par 
For two points of the deflection field $\boldsymbol{x}_{w,i}$ and $\boldsymbol{x}_{w,j}$, the exponential ARD kernel yields:
\begin{equation}\label{eq:Kernel}
	k_{ww}(\boldsymbol{x}_{w,i},\boldsymbol{x}_{w,j}^\prime;\boldsymbol{\theta})=A^2\exp\left(-\frac{1}{2}\frac{(x_{w,i}-x_{w,j})^2}{l_x^2}-\frac{1}{2}\frac{(y_{w,i}-y_{w,j})^2}{l_y^2}\right),
\end{equation}
where $\boldsymbol{\theta}=(A,l_x,l_y)^T$ is the vector containing the kernel hyperparameters. This vector includes the variance $A^2$ and the characteristic length-scales $l_x$ and $l_y$ for each direction, which represent the anisotropicity. The length-scales and variance for each model are not known \textit{a priori}. They are inferred based on the available data using Bayesian inference, as discussed in the following section. The exponential kernel has shown empirical success for the present application. However, the question of which kernel is most suitable for applications warrants further investigation.\par
The prior given in~\eqref{eq:MultiOutput_GP_PlateShort}, \eqref{eq:MultiOutput_GP_Plate} encodes both the statistical assumptions through the displacement covariance $k_{ww}$ and mechanical assumptions through the other cross-covariance functions. Typically, the true values $\boldsymbol{f}$ are not directly available from the measurement data. Instead, we have access to their noisy observations $\boldsymbol{z}$, which we model as
\begin{equation}\label{eq:Decomp}
	\boldsymbol{z}=\boldsymbol{f}+\boldsymbol{e},
\end{equation}
where $\boldsymbol{e}\in{\mathbb{R}^{N\times1}}$ is measurement noise.
We assume independent identically distributed Gaussian noise:
\begin{equation}
	\boldsymbol{e}\sim p(\boldsymbol{e}|\boldsymbol{\sigma}^2)=\mathcal{N}(\boldsymbol{0},\boldsymbol{E}).
\end{equation}
Here, $\boldsymbol{E}\in{\mathbb{R}}^{N\times N}$ is a diagonal block unit matrix, which block members on the diagonal contain the noise variance $\sigma^2$ of the corresponding physical quantity. Example for the case of the deflection, the noise matrix yields $\boldsymbol{E}_w=\sigma_w^2\boldsymbol{I}_w$ for $\boldsymbol{I}_w\in\mathbb{R}^{N_w\times N_w}$  being an identity matrix. We consolidate the variances of individual physical quantities in a vector $\boldsymbol{\sigma}^2$. Since the prior $\boldsymbol{f}$ and measurement noise $\boldsymbol{e}$ are Gaussian in~\eqref{eq:Decomp}, the \textit{likelihood} of the data $\boldsymbol{z}$ given the priors $\boldsymbol{f}$ and $\boldsymbol{e}$ is
\begin{equation}\label{eq:likelihood}
 p(\boldsymbol{z}|\boldsymbol{f},\boldsymbol{X},\boldsymbol{\theta}^e)=\mathcal{N}(\boldsymbol{f},\boldsymbol{E}),
\end{equation}
where the extended hyperparameter vector $\boldsymbol{\theta}^e$ includes the flexural rigidity and noise variances:
\begin{equation}
	\boldsymbol{\theta}^e=\left(\boldsymbol{\theta},D,\boldsymbol{\sigma}^2\right)^T.
\end{equation}
Finally, the boundary conditions are considered by using the simple method of artificial observations. This involves restricting the function values at locations $\boldsymbol{X}_{\mathrm{BC}}$ by introducing noiseless observations depending on the boundary conditions in~\eqref{eq:BC_C} as:
\begin{equation}\label{eq:BC}
	\boldsymbol{z}_{\mathrm{BC}}=\boldsymbol{z}(\boldsymbol{X}_\mathrm{BC})=\boldsymbol{f}_\mathrm{BC}=0,
\end{equation}
which implies zero measurement noise $ \boldsymbol{E}(\boldsymbol{X}_\mathrm{BC},\boldsymbol{X}_\mathrm{BC})=0.$
Although there are more elegant way of applying boundary conditions through the Gaussian prior~\cite{gulianGaussianProcessRegression2022}, enforcing boundary conditions using artificial noiseless observations way is straightforward. However, it may affect numerical stability in the inverse of the covariance matrix.\par 
Figure~\ref{fig:Schematic_PlateTheory} depicts the physics-informed model schematically. The prior~\eqref{eq:MultiOutput_GP_PlateShort} and likelihood~\eqref{eq:likelihood} are the key descriptors of the presented physics-informed statistical model. In the what follows, these are used for Bayesian inference of the \textit{posteriors} of the hyperparameters (including the rigidity $D$) and predictions through marginalization of these two terms based on measurement data.\par

\subsection{Learning}\label{sec:Learning}
Simultaneous learning of the extended hyperparameters $\boldsymbol{\theta}^e$ delivers the noise variance, kernel hyperparameters (variance, length-scales), and flexural rigidity, thereby enabling physics-informed learning. These parameters are not known \textit{a priori}, and are estimated from the data in $\boldsymbol{z}$, which includes response measurements and boundary conditions. The data can be any of the plate physical quantities; however, the covariance matrix $\boldsymbol{K}$ must consist of at least two block terms if the rigidity is inferred: one excluding and one including the rigidity $D$. Learning the rigidity $D$ based on heterogeneous noisy data is one of the key goals of the presented model. In practical terms, internal forces are not measured directly and therefore cannot be used to infer rigidity. They can only be inferred from the data and are applicable only for prediction after the learning process is complete, as will be discussed in the following section.  \par

We use two methods to infer $\boldsymbol{\theta}^e$: a) by providing estimates through maximizing the log marginal likelihood, and b) by providing full distribution through MCMC sampling. The members of $\boldsymbol{\theta}^e$ are maximum likelihood estimates (MLE) $\hat{\boldsymbol{\theta}^e}$ for the first method, while probability distributions $p(\boldsymbol{\theta}^e|\boldsymbol{z},\boldsymbol{X})$ for the second method. \par
Using Bayes' theorem, the \textit{parameter posterior} is inferred as 
\begin{equation}\label{eq:Lvl2Inference}
	p(\boldsymbol{\theta}^e|\boldsymbol{z},\boldsymbol{X})=\frac{p(\boldsymbol{z}|\boldsymbol{X},\boldsymbol{\theta}^e)p(\boldsymbol{\theta}^e)}{p(\boldsymbol{z}|\boldsymbol{X})}
	\propto p(\boldsymbol{z}|\boldsymbol{X},\boldsymbol{\theta}^e)p(\boldsymbol{\theta}^e),
\end{equation}
where the $p(\boldsymbol{\theta}^e)$ is the \textit{hyperprior}, $p(\boldsymbol{z}|\boldsymbol{X},\boldsymbol{\theta}^e)$ is the \textit{marginal likelihood} (i.e. evidence), and $p(\boldsymbol{z}|\boldsymbol{X})$ is a \textit{normalization constant} w.r.t. $\boldsymbol{\theta}^e$. The latter can be omitted for the posterior $p(\boldsymbol{\theta}^e|\boldsymbol{z},\boldsymbol{X})$ only if it is sampled through MCMC or if MLE estimates $\hat{\boldsymbol{\theta}^e}$ are computed, as in our case. The marginal likelihood $p(\boldsymbol{z}|\boldsymbol{X},\boldsymbol{\theta}^e)$ is obtained by marginalizing the function values $\boldsymbol{f}$ out of the product of the likelihood~\eqref{eq:likelihood} and the prior~\eqref{eq:MultiOutput_GP_PlateShort}:
\begin{equation}\label{eq:MarginalLikelihood}
	p(\boldsymbol{z}|\boldsymbol{X},\boldsymbol{\theta}^e)=\int p(\boldsymbol{z}|\boldsymbol{f},\boldsymbol{X},\boldsymbol{\theta}^e)p(\boldsymbol{f}|\boldsymbol{X},\boldsymbol{\theta}^e)\mathrm{d}\boldsymbol{f}.
\end{equation}
Working in a logarithmic form,~\eqref{eq:Lvl2Inference} becomes a sum
\begin{equation}\label{eq:LogLvl2Inference}
	\log p(\boldsymbol{\theta}^e|\boldsymbol{z},\boldsymbol{X}) \propto \log p(\boldsymbol{z}|\boldsymbol{X,\boldsymbol{\theta}^e)}+\log p(\boldsymbol{\theta}^e),
\end{equation}
where the log marginal likelihood $\log p(\boldsymbol{z}|\boldsymbol{X,\boldsymbol{\theta}^e)}$ in~\eqref{eq:MarginalLikelihood} is analytically tractable based on the likelihood~\eqref{eq:likelihood} and prior~\eqref{eq:MultiOutput_GP_PlateShort}, yielding:
\begin{equation}\label{eq:LogMargLik}
 \log p(\boldsymbol{z}|\boldsymbol{X},\boldsymbol{\theta}^e)=-\frac{1}{2}\boldsymbol{z}^T(\boldsymbol{\boldsymbol{K}+\boldsymbol{E}})^{-1}\boldsymbol{z}-\frac{1}{2}\log|\boldsymbol{K}+\boldsymbol{E}|-\frac{N}{2}\log2\pi.
\end{equation}\par
The MLE estimate $\hat{\boldsymbol{\theta}^e}$ (method a) is obtained by maximizing~\eqref{eq:LogMargLik}:
\begin{equation}\label{eq:MLE}
	\hat{\boldsymbol{\theta}^e}=\arg \max_{\boldsymbol{\theta}^e} \log p(\boldsymbol{z}|\boldsymbol{X},\boldsymbol{\theta}^e),
\end{equation}
which is equivalent to the maximum value of the log posterior $	p(\boldsymbol{\theta}^e|\boldsymbol{z},\boldsymbol{X})$, i.e. maximum a posteriori estimate (MAP) in~\eqref{eq:Lvl2Inference} for a uniform flat hyperprior $p(\boldsymbol{\theta}^e)$ (i.e. $\log p(\boldsymbol{\theta}^e)=const$). We employ the conjugate gradient decent optimizer~\cite{rasmussenGaussianProcessesMachine2006} for solving~\eqref{eq:MLE}, by supplying the partial derivatives:

\begin{equation}\label{eq:LikelihoodDer}
	\begin{aligned}
		\frac{\partial}{\partial\theta_{j}^e} \log p(\boldsymbol{z}|\boldsymbol{X},\boldsymbol{\theta}^e)=\frac{1}{2}\mathrm{tr}\Bigg(\boldsymbol{r}\boldsymbol{r}^T\frac{\partial(\boldsymbol{K}+\boldsymbol{E}) }{\partial\theta_{j}^e}-(\boldsymbol{K}+\boldsymbol{E})^{-1}\frac{\partial(\boldsymbol{K}+\boldsymbol{E}) }{\partial\theta_{j}^e}\Bigg),
	\end{aligned}
\end{equation}

where $\boldsymbol{r}=(\boldsymbol{K}+\boldsymbol{E})^{-1}\boldsymbol{z}$ and $\mathrm{tr}(\cdot)$ is the trace.\par
The full probabilistic distribution $p(\boldsymbol{\theta}^e|\boldsymbol{z},\boldsymbol{X})$ (method b) is obtained by MCMC sampling of~\eqref{eq:LogLvl2Inference} using the Metropolis–Hastings (MH) algorithm~\cite{hastingsMonteCarloSampling1970} (see Appendix~\ref{App:MH}), based on~\eqref{eq:LogMargLik} and assumed uniform hyperprior $p(\boldsymbol{\theta}^e)$. For comparison with the MLE estimate, it is useful to define the mean of the sampled distribution as:\par
\begin{equation}\label{eq:MCMC}
	\overline{{\boldsymbol{\theta}}^e}=\frac{1}{N_s}\sum_{i=1}^{N_s}\boldsymbol{\theta}^e_i,
\end{equation}
where $\boldsymbol{\theta}^e_i\sim p(\boldsymbol{\theta}^e|\boldsymbol{z},\boldsymbol{X})$ are $N_s$ number of draws from the posterior of the hyperparameters based on the MH algorithm.\par

The full probability distribution of the extended hyperparameters obtained from MCMC sampling provides a more comprehensive representation of the uncertainty compared to the MLE estimate. In contrast, obtaining the MLE estimate is computationally more efficient since it involves using a gradient solver on a convex function. However, it may be hindered as $p(\boldsymbol{z}|\boldsymbol{X},\boldsymbol{\theta}^e)$ may have multiple local minima. These two methods are further discussed for the present application in Section~\ref{Sec:NumExp}. Having noiseless boundary conditions (cf.~\eqref{eq:BC}) impacts the numerical stability when inverting the kernel and noise matrices in~\eqref{eq:LogMargLik}. This requires adding a jitter $\varepsilon$ (artificial numerical noise, typically up to 1e-5) to the diagonal of the covariance matrix to ensure numerical stability when inverting it.\par 
From a practical aspect, it makes sense to construct $\boldsymbol{K}$ during learning based on physical quantities that are easier to measure from sensors (e.g. deflection and load) than others (e.g. internal moments); however, we retain the generality as the data can also come from numerical experiments.

\subsection{Prediction}\label{sec:Prediction}
Once the hyperparameters (MLE estimates or full distributions) are obtained, the model can be utilized to predict the physical quantities of the plate at the prediction points $\boldsymbol{X}^*$. The model has the capability to predict both observed and unobserved physical quantities, with observed quantities being those for which measurements exist at any point(s) in the plate. \par
The \textit{predictive posterior} of the function values $\boldsymbol{f}^*$ is obtained based on the joint distribution  $p(\boldsymbol{f}^*,\boldsymbol{\theta}^e|\boldsymbol{X}^*,\boldsymbol{z},\boldsymbol{X})$ at prediction points $\boldsymbol{X}^*$, given the data $\mathcal{D}$. Marginalizing the joint distribution over the hyperparameters using the chain rule yields
\begin{equation}~\label{eq:PredDistFull}
	\begin{aligned}
p(\boldsymbol{f}^*|\boldsymbol{X}^*,\boldsymbol{z},\boldsymbol{X})&=\int p(\boldsymbol{f}^*,\boldsymbol{\theta}^e|\boldsymbol{X}^*,\boldsymbol{z},\boldsymbol{X})\mathrm{d}\boldsymbol{\theta}^e\\
&=
\int p(\boldsymbol{f}^*|\boldsymbol{\theta}^e,\boldsymbol{X}^*,\boldsymbol{z},\boldsymbol{X})p(\boldsymbol{\theta}^e|\boldsymbol{z},\boldsymbol{X})\mathrm{d}\boldsymbol{\theta}^e.
\end{aligned}
\end{equation}
The second member in the preceding relation is hyperparameter distribution from~\eqref{eq:Lvl2Inference}, while the first member is the standard predictive GP posterior conditioned on the hyperparameters. To obtain the first term, first consider the predictions $\boldsymbol{f}^*$ and noisy observations $\boldsymbol{z}$ as a joint multivariate Gaussian:
\begin{equation}~\label{eq:Joint}
\begin{bmatrix}
	\boldsymbol{z}\\
	\boldsymbol{f}^*
\end{bmatrix}\sim 
\mathcal{N}\left(\boldsymbol{0},\begin{bmatrix}
	\boldsymbol{K}+\boldsymbol{E} & \boldsymbol{K}_*\\
\boldsymbol{K}_*^\prime & \boldsymbol{K}_{**}
\end{bmatrix}\right),
\end{equation}
where the covariance matrices, dependent on the prediction points $\boldsymbol{X}^*$, are: $\boldsymbol{K}_{**}=\boldsymbol{\mathcal{K}}(\boldsymbol{X}^*,\boldsymbol{X}^*)\in\mathbb{R}^{N^*\times N^*}$, $\boldsymbol{K}_{*}=\boldsymbol{\mathcal{K}}(\boldsymbol{X},\boldsymbol{X}^*)\in\mathbb{R}^{N\times N^*}$ and $\boldsymbol{K}_{*}^\prime=\boldsymbol{\mathcal{K}}(\boldsymbol{X}^*,\boldsymbol{X})\in\mathbb{R}^{N^*\times N}$ (for convenience, the dependence on $\boldsymbol{\theta}^e$ is dropped for the covariance matrices). Conditioning the joint distribution~\eqref{eq:Joint} on the observations, the marginal predictive posterior is analytically tractable~(see \cite{murphyMachineLearningProbabilistic2012}):
\begin{equation}~\label{eq:PredDist}
%	\begin{aligned}
	p(\boldsymbol{f}^*|\boldsymbol{\theta}^e,\boldsymbol{X}^*,\boldsymbol{z},\boldsymbol{X})
%	&=\int p(\boldsymbol{f}^*,\boldsymbol{f}|\boldsymbol{\theta}^e,\boldsymbol{X}^*,\boldsymbol{z},\boldsymbol{X})\mathrm{d}\boldsymbol{f}\\
%	&=\int p(\boldsymbol{f}^*|\boldsymbol{f},\boldsymbol{\theta}^e,\boldsymbol{X}^*)p(\boldsymbol{f}|\boldsymbol{\theta}^e,\boldsymbol{z},\boldsymbol{X})\mathrm{d}\boldsymbol{f}\\
	=\mathcal{N}(\boldsymbol{m}^*,\boldsymbol{K}^*)
%\end{aligned}
\end{equation}
where the predictive mean $\boldsymbol{m}^*\in\mathbb{R}^{N^* \times 1}$ and covariance $\boldsymbol{K}\in\mathbb{R}^{N^* \times N^*}$ are
\begin{equation}~\label{eq:PredictiveMean}
\begin{aligned}
	\boldsymbol{m}^*&={\boldsymbol{K}_*^T}\left(\boldsymbol{K}+\boldsymbol{E}\right)^{-1}\boldsymbol{z},\\
	\boldsymbol{K}^*&=\boldsymbol{K}_{**}-\boldsymbol{K}_*^T\left(\boldsymbol{K}+\boldsymbol{E}\right)^{-1}\boldsymbol{K}_*.
\end{aligned}
\end{equation}
If the MLE estimate $\hat{\boldsymbol{\theta}^e}$ is used, then the predictive posterior in~\eqref{eq:PredDistFull} reduces to~\eqref{eq:PredDist} with a constant $\hat{\boldsymbol{\theta}^e}$. If full distributions are used, as in our case, then the posterior in~\eqref{eq:PredDistFull} is analytically intracable. Therefore, it is estimated using Monte Carlo sampling as
\begin{equation}~\label{eq:Sampled}
		p(\boldsymbol{f}^*|\boldsymbol{X}^*,\boldsymbol{z},\boldsymbol{X})\approx \frac{1}{N_s}\sum_{i=1}^{N_s}p(\boldsymbol{f}^*|\boldsymbol{\theta}^e_i,\boldsymbol{X}^*,\boldsymbol{z},\boldsymbol{X})
\end{equation}
where $\boldsymbol{\theta}^e_i\sim p(\boldsymbol{\theta}^e|\boldsymbol{z},\boldsymbol{X})$ are draws from the distribution and $N_s$ is the number of number of draws.\par
The predictive quantities $\boldsymbol{f}^*$ at $\boldsymbol{X}^*$ do not need to be measured anywhere on the plate, rather they can be inferred based on observations of other physical quantities. Example, a model trained based on deflection and load observations $\boldsymbol{z}_{wq}$ can infer the moment $M_y$, i.e. $\boldsymbol{f}_{M_y}^*$. In this case, the block covariance matrix $\boldsymbol{K}$ at the observation is comprised of $\boldsymbol{K}_{ww}=k_{ww}(\boldsymbol{X}_{w},\boldsymbol{X}_{w})$, $\boldsymbol{K}_{wq}^D=k_{wq}^D(\boldsymbol{X}_{w},\boldsymbol{X}_{q})$, $\boldsymbol{K}_{qw}^D=k_{qw}^D(\boldsymbol{X}_{q},\boldsymbol{X}_{w})$ and $\boldsymbol{K}_{qq}^D=k_{qq}^D(\boldsymbol{X}_{q},\boldsymbol{X}_{q})$; while the predictive covariance matrix is $\boldsymbol{K}_{**}=k_{M_y}^D(\boldsymbol{X}_{M_y}^*,\boldsymbol{X}_{M_y}^*)$. The cross matrices are obtained based on the cross covariance functions between the observation and predictive quantities as $\boldsymbol{K}_*=\left[k_{wM_y}^D(\boldsymbol{X}_w,\boldsymbol{X}_{M_y}^*);k_{qM_y}^D(\boldsymbol{X}_q,\boldsymbol{X}_{M_y}^*)\right]$ and $\boldsymbol{K}_*^\prime=\left[k_{M_yw}^D(\boldsymbol{X}_{M_y}^*,\boldsymbol{X}_{w});k_{M_yq}^D(\boldsymbol{X}_{M_y}^*,\boldsymbol{X}_{q})\right]$. It is, however, noted that if the internal forces are to be inferred based on other physical quantities, the Poisson ratio needs to be known \textit{a priori}. Alternatively, in the less likely situation where the Young's modulus and thickness are known, the Poisson ratio can be inferred from the structural rigidity.

\begin{figure}[!b]
	\centering
	\includegraphics[clip]{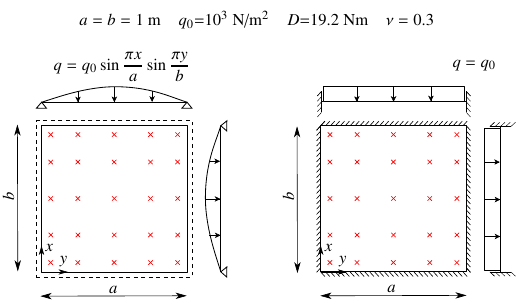} 
	\caption{Numerical experiments: i) simply-supported plate subjected to sinusoidal load (left); ii) fixed plate subjected to uniform load (right). The red crosses represent the observation locations $\boldsymbol{X}$.}
	\label{fig:Schematic_PlateEx}
\end{figure} 
\section{Numerical Experiments}\label{Sec:NumExp}
In this section, we conduct numerical experiments for two examples with analytical solutions: i) a simply supported plate subjected to sinusoidal loading, and ii) a fixed plate subjected to uniform loading. The two examples and the corresponding parameters are depicted in Fig.~\ref{fig:Schematic_PlateEx}. The training data is available on 5 observation locations in each direction for the learning (i.e. a total of 25 points), while the prediction is at 21 locations in each direction (i.e. a total of 441 points). The training locations are equidistant, except at the boundaries, where the last points are moved 5\% inward (see Fig.~\ref{fig:Schematic_PlateEx}). Unless otherwise noted, noise with a signal-to-noise (SNR) ratio of 10 is added to the analytical training data to account for noisy observations. The two objectives of the experiments are the same as previously, i.e., learning of the flexural rigidity $D$ and prediction of physical plate quantities at unobserved locations.\par
The model is implemented in Matlab, using symbolic computation toolbox to obtain the covariance functions derived in Appendix~\ref{App:CovFunc}. We use Cholesky decomposition for the inversion of the covariance matrix for computational efficiency during learning and prediction. The implementation for the conjugate gradient optimizer of~\cite{rasmussenGaussianProcessesMachine2006}~is used for the MLE estimation. The code is available in an open-source repository: \href{https://github.com/IgorKavrakov/PlateGP}{github.com/IgorKavrakov/PlateGP}.

\begin{figure}[!t]
	\centering
	\includegraphics[clip]{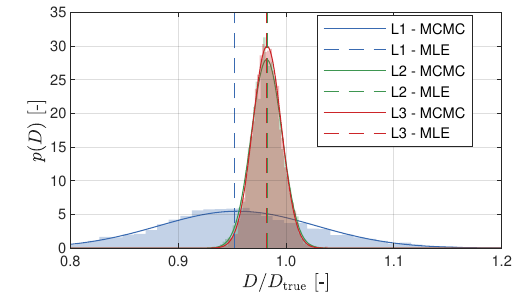} 
	\caption{Simply-supported plate with sinusoidal loading: Histogram of MCMC samples for the flextural rigidity based on a single set of measurements ($N_o$=1). The dashed lines represent the MLE estimates.}
	\label{fig:Ex1a_Histogram}
\end{figure}

\begin{figure}[!t]
	\centering
	\includegraphics[clip]{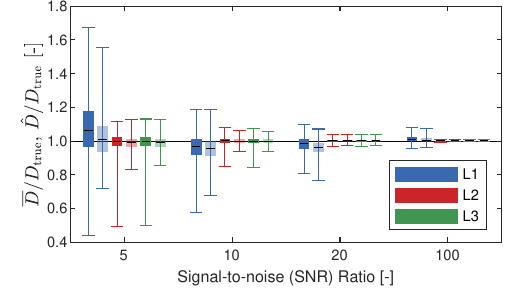} 
	\caption{Simply-supported plate with sinusoidal loading: Monte Carlo analysis for $N_o$=1000 sets of measurements for each signal-to-noise ratio. Mean of MCMC means $\overline{D}$ (opaque; cf.~\eqref{eq:MCMC}) and the MLE estimates $\hat{D}$ (transparent; cf.~\eqref{eq:MLE}) of the learned flexural rigidities, and their corresponding 25\% / 75\% quantiles and minimum/maximum values.}
	\label{fig:Ex1b_Histogram}
\end{figure} 

\begin{figure}[!t]
	\centering
	\includegraphics[clip]{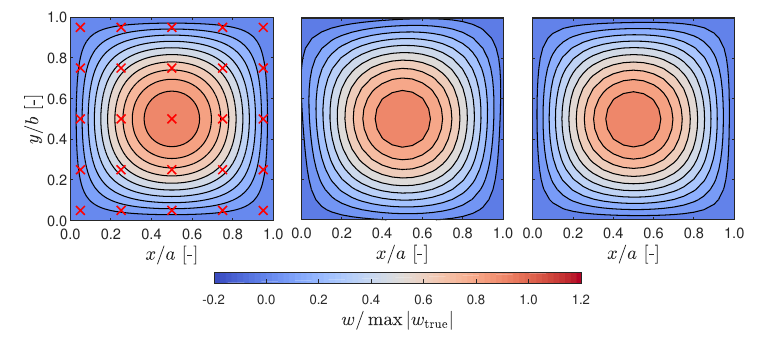} 
	\includegraphics[clip]{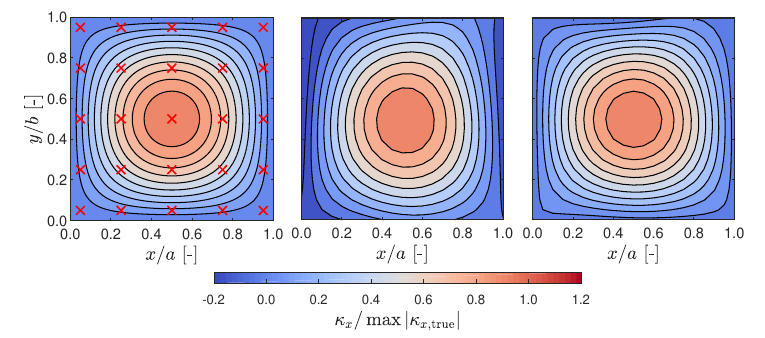} 
	\includegraphics[clip]{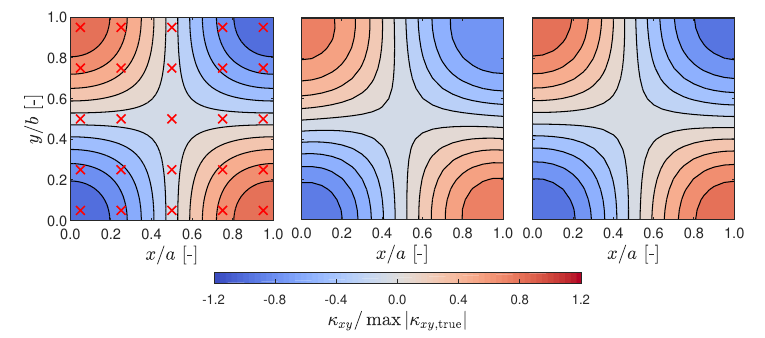} 
	\caption{Simply-supported plate with sinusoidal loading: Normalised mean deflection $w$ (top), curvatures $\kappa_x$ (centre) and $\kappa_{xy}$ (bottom) - True values (left), prediction for L1 (center) and prediction for L3 (right). The red crosses represent the training points for the corresponding learning cases.}
	\label{fig:Ex1a_Contour_w}
\end{figure} 
\subsection{Simply-supported plate subjected to sinusoidal load}
%\subsubsection{Analytical solution}
The exact solution of the deflection for the simply-supported plate with sinusoidal load is~\cite{timoshenkoTheoryPlatesShells1987}:
\begin{equation}
 w=\frac{q_0}{\pi^4D\left(\displaystyle\frac{1}{a^2}+\frac{1}{b^2}\right)}\sin\frac{\pi x}{a}\sin\frac{\pi y}{b}.
\end{equation}
The other physical quantities (rotations, curvatures and internal forces) can be obtained directly based on the deflection using the linear operators in~\eqref{eq:Operators}.\par
First, we study how well the flexural rigidity parameter $D$ can be identified from noisy observations, depending on the SNR ratio and measurement type. The type of data considered for learning are the deflection $w$, curvatures $\kappa_x,\kappa_y,\kappa_{xy}$ and the load $q$. Three learning cases are considered based on the formulation of the observation data set: 
\begin{itemize}
	\item[L1:] Learning based on the load $q$ and deflection $w$ - observation data set $\mathcal{D}_{wq}=(\boldsymbol{X}_{wq},\boldsymbol{z}_{wq})$,
	\item [L2:] Learning based on the load $q$ and curvatures $\kappa_x,\kappa_y,\kappa_{xy}$ - observation data set $\mathcal{D}_{\kappa q}=(\boldsymbol{X}_{\kappa q},\boldsymbol{z}_{\kappa q})$,
	\item [L3:] Learning based on the load $q$, curvatures $\kappa_x,\kappa_y,\kappa_{xy}$ and deflection $w$ - observation data set $\mathcal{D}_{w\kappa q}=(\boldsymbol{X}_{w\kappa q},\boldsymbol{z}_{w\kappa q})$.
\end{itemize}

\begin{figure}[!t]
	\centering
	\includegraphics[clip]{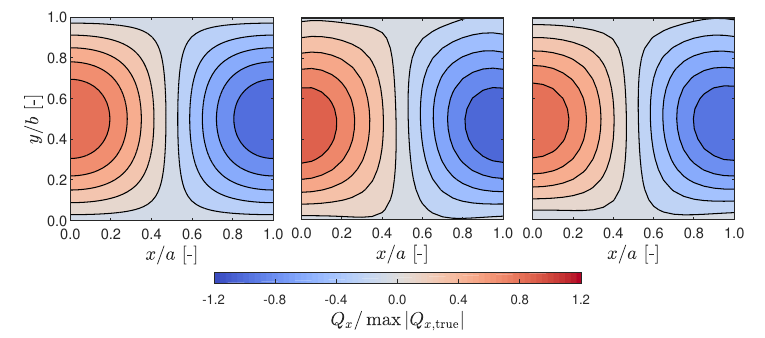} 
	\includegraphics[clip]{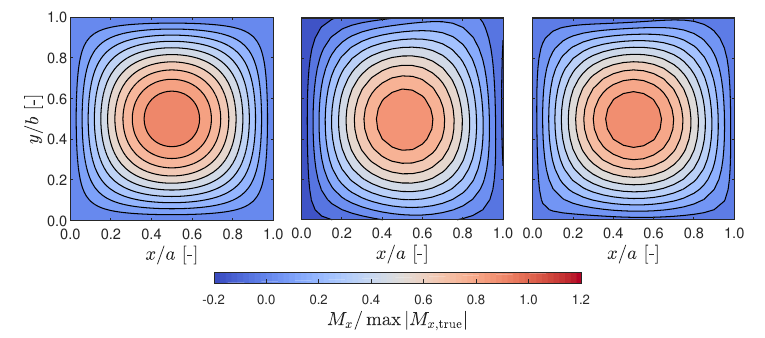} 
	\includegraphics[clip]{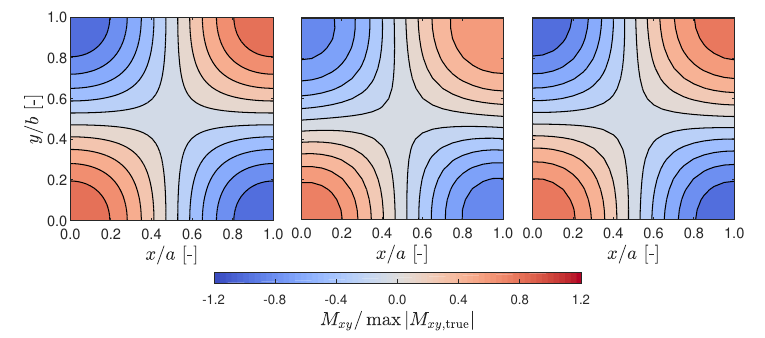} 
	\caption{Simply-supported plate with sinusoidal loading: Normalised mean shear force $Q_x$ (top), moments $M_x$ (centre) and $M_{xy}$ (bottom) - True values (left), prediction for L1 (center) and prediction for L3 (right).}
	\label{fig:Ex1a_Contour_Int}
\end{figure}

\begin{figure}[!t]
	\centering
	\hspace*{-1.2cm}\includegraphics[clip,scale=1.0]{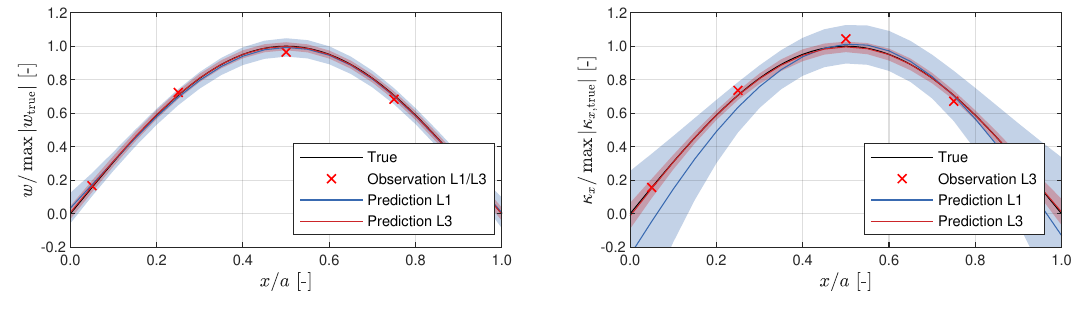}\vspace*{-0.3cm}
	\hspace*{-1.2cm}\includegraphics[clip,scale=1.0]{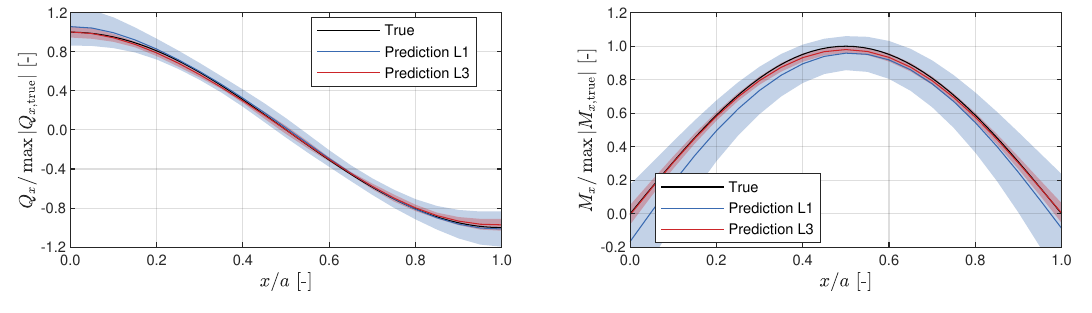}\vspace*{-0.3cm}
	\hspace*{-1.2cm}\includegraphics[clip,scale=1.0]{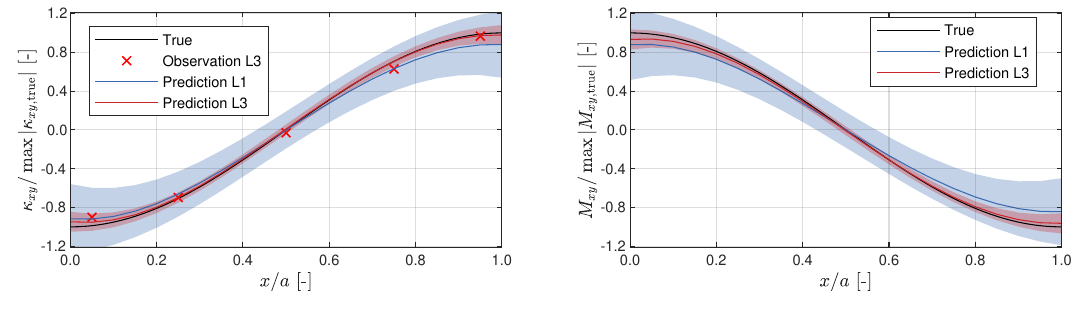}
    %\hspace*{-1.2cm}\includegraphics[clip,scale=1.0]{Figures//Combined_Figure7//Ex1a_CenterLine_wKx_QxMx_KxyMxy.pdf}
    \caption{Simply-supported plate with sinusoidal loading: deflection $w$ (top-left), curvature $\kappa_x$ (top-right), shear force $Q_x$ (centre-left), moment $M_x$ (centre-right), curvature $k_{xy}$ (bottom-left) and moment $M_{xy}$ (bottom-right) at the centreline $y=0.5b$ (see Figs.~\ref{fig:Ex1a_Contour_w} and~\ref{fig:Ex1a_Contour_Int}) for the cases L1 and L2. The true values are the analytical solution; the observations (red crosses), used for training, include the true values and additional measurement noise; the prediction is obtained using the GP model. The shaded areas in the corresponding colours are the 99\% confidence interval for the prediction.}
	\label{fig:Ex1a_CenterLine_wKx}
\end{figure}

All of the learning cases use load-dependent covariance functions to encode the flexural rigidity $D$. L1 is designed to maximize the influence of the statistical assumptions in the covariance $k_{ww}$. In the second case, L2, the derivatives of $k_{ww}$ are included, which incorporates more physical knowledge about the system. The key point is that although the curvatures are observed independently, they are related through their cross-covariances due to their relationship with the deflection, despite the absence of deflection observations. This is an advantage of using the proposed model since, otherwise, the curvatures would have been simply superimposed if a simple model updating procedure was used. Since all three curvatures are considered separately, there is more data plugged into the model. L3 is constructed to demonstrate how several types of physical quantities can be integrated into the learning process. \par
Figure~\ref{fig:Ex1a_Histogram} depicts the learned rigidity for a single set ($N_o$ = 1) of observation data with SNR=10 using both MLE and MCMC.  The full probability $p(D)$ is depicted for the MCMC learning, while MLE results in point estimates. The MLE estimates mostly coincide with the peak of the MCMC distributions; however, it should be noted that this may not always be the case if there is a presence of non-Gaussian distributions or if there are numerical issues in the gradient optimizer. In both cases, the values obtained are relatively close to the true value, taking into consideration that the observation data is contaminated by noise. The L1 case resulted in distribution with larger spread and generally worse prediction (error for the mean is 4.5\%) than the L2 and L3 cases (error for the mean is 1.8\%). It is interesting to observe that supplying the curvatures instead of the deflections results in better estimates of the flexural rigidity. Two reasons can explain this behavior. First, because L2 has a larger learning data set (three curvatures vs one deflection) and second, because supplying the regression derivatives (i.e. curvatures) contains more information about the system than the actual regression values (i.e. deflections)~\cite{solakDerivativeObservationsGaussian2002,sarkkaLinearOperatorsStochastic2011}. The L3 case resulted in the best estimate with the smallest spread of the distribution. This is expected as the complete data related through mechanical laws in the GP model is used for learning. Nevertheless, the results do not deviate much from the L2 case, meaning that the curvature data governs the learning. \par
To obtain statistical significance of the results, we conducted a Monte Carlo analysis by re-sampling the noise for the analytical true data. This resulted in $N_o=1000$ different sets of observations for four SNR ratios: $\mathrm{SNR}=\sigma_{\mathrm{data}}/\sigma_{\mathrm{noise}}=\lbrace5,10,20,100\rbrace$. The flexural rigidity is learned based on both MLE and MCMC strategies with identical initial values of $\boldsymbol{\theta}^*$. Figure~\ref{fig:Ex1b_Histogram} shows the mean of 1000 MCMC means $\overline{D}$ (opaque; cf.~\eqref{eq:MCMC}) and the MLE estimates $\hat{D}$ (transparent; cf.~\eqref{eq:MLE}) of the learned flexural rigidities, their corresponding 25\% / 75\% quantiles and the corresponding minimum/maximum values. We make several points based on these results. First, the L2 and L3 cases consistently showed better prediction, which confirms the findings for the single set of observations we previously examined (Fig.~\ref{fig:Ex1a_Histogram}). Second, higher SNR ratios, and therefore less noise content in the data, result in better predictions—both in terms of a more accurate mean and a smaller confidence interval. This is a consequence of Occam's razor, which balances between data fit and model complexity when fitting the GP regression to the data. Thus, measurement uncertainty propagates through the model and influences the stiffness estimation. However, it is expected that the mean for all learning cases converges to the true value for larger number of sets of observations $N_o$. Third, the MCMC mean estimate consistently shows better results with lower confidence interval than the MLE estimate. Moreover, despite being computationally faster, the MLE estimation may result in wrong estimates (e.g. $\hat{D}\leq 0.2D_{\mathrm{true}})$ since the gradient optimizer can get stuck in a local minimum (this is also observed to a certain extent in the minimum/maximum values). Although this is usually avoidable by re-initiating with modified the initial values of $\boldsymbol{\theta}^*$; herein, these MLE estimates were disregarded since their number is insignificant - up to 1.8\% of the 1000 sets of observations.\par
Having learned the flexural rigidity and the hyperparameters based on L1 and L3  (cf. Fig~\ref{fig:Ex1a_Histogram}), we use the MCMC to predict observed and unobserved quantities at unobserved locations. Figure~\ref{fig:Ex1a_Contour_w} shows the mean prediction of the deflections $w$ and curvatures $\kappa_x$ and $\kappa_{xy}$, while Fig.~\ref{fig:Ex1a_Contour_Int} shows the shear force $Q_x$ and moments $M_x$ and $M_{xy}$. Further, Fig.~\ref{fig:Ex1a_CenterLine_wKx} shows the mean and 99\% confidence interval of all these quantities at the centreline $y=0.5b$. In all figures, the prediction of the physics-informed GP model is normalized w.r.t. the maximum analytical "true" values. The mean prediction for both learning cases resulted in similar values for the deflections (cf. Fig.~\ref{fig:Ex1a_Contour_w} top; Fig.~\ref{fig:Ex1a_CenterLine_wKx} top-left), with higher uncertainty for the L1 case. However; the mean prediction of the curvatures (cf. Fig.~\ref{fig:Ex1a_Contour_w} center and bottom; Fig.~\ref{fig:Ex1a_CenterLine_wKx} top-right and bottom-left) is significantly better for the L3 case with lower uncertainty. The internal forces (cf. Fig.~\ref{fig:Ex1a_Contour_Int}; Fig.~\ref{fig:Ex1a_CenterLine_wKx} top-right and bottom-left) also show similar behavior; however, the difference here is that the internal forces are unobserved quantities for both L1 and L3 cases (i.e. no training data). The reason for the better prediction in case of L3 learning is obvious - there is additional training data that helps in the conditioning for the learning and prediction. \par
Overall, the results for all learning cases demonstrate that the proposed methodology is effective in predicting both observed and unobserved physical quantities. The differences between the mean predictions and the analytical true solutions are anticipated, as they arise from the observation noise. However, it is important to note that the observations and the true analytical values lie within the prediction's confidence interval.

\begin{figure}[!b]
	\centering
	\includegraphics[clip]{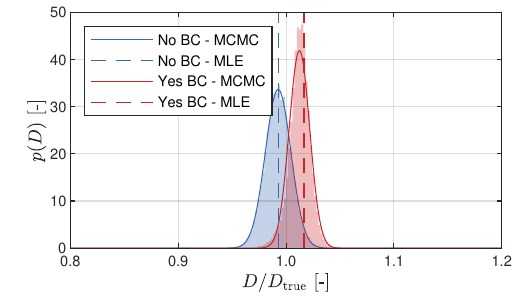} 
	\caption{Fixed plate with uniform loading: Histogram of MCMC samples for the flextural rigidity for the cases with and without boundary conditions (BC). The dashed lines represent the MLE estimates.}
	\label{fig:Ex2_Histogram_D_MCMC}
\end{figure} 

\begin{figure}[!b]
	\centering
	\includegraphics[clip,]{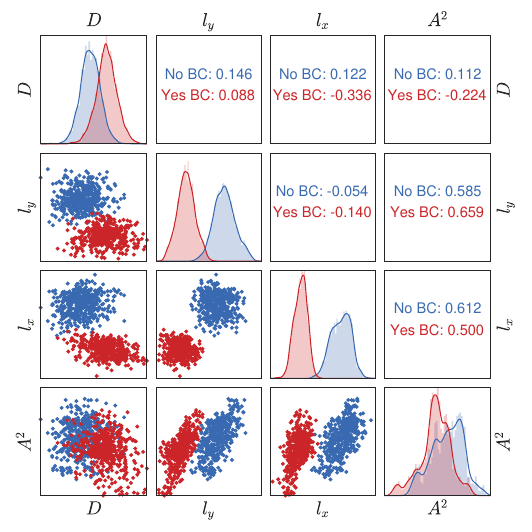} 
	\caption{Fixed plate with uniform loading: Correlation plots and MCMC histograms of the kernel hyperparameters ($l_y,l_x,A$) and the flexural rigidity ($D$) for the cases with and without boundary conditions (BC). The upper triangle shows the correlation coefficients.}
	\label{fig:Ex2_Parameters}
\end{figure}

\subsection{Fixed plate subjected to uniform load}
The second example is a fixed plate subjected to uniform load (see Fig.~\ref{fig:Schematic_PlateEx}, right). For this case, there is no exact analytical solution for the deflection and there are several approximate solution methods~\cite{meleshkoBendingElasticRectangular1997}. Here, we use the double-cosine solution of Taylor and Govindjee~\cite{taylorSolutionClampedRectangular2004}:
\begin{equation}
w(\boldsymbol{x})=\sum_{m=1}^{N_m}\sum_{n=1}^{N_n}\left(1-\cos\frac{2m\pi x}{a}
\right)\left(1-\cos\frac{2n\pi y}{b}\right)w_{mn},
\end{equation}
where $w_{mn}$ are parameters that are determined using the Ritz method on~\eqref{eq:EqPlate} and $N_m$ and $N_n$ are Fourier coefficients, which are taken as $N_n=N_m=200$. The curvatures can be obtained using the operators in~\eqref{eq:Curvature}. The analytical data for learning and prediction is similar as in the previous example of simply supported plate. Here, we use only the L3 case for learning, i.e. the training data contains values of the deflections, curvature and loading, for a single set of observation data  with SNR=10. However, the effect of the boundary conditions on the displacement and rotation fields~\eqref{eq:BC_C} is studied by imposing them through noiseless observations as in~\eqref{eq:BC}. Curvatures could also serve as boundary conditions for the plate model, potentially enhancing inference accuracy, though this might compromise numerical stability when numerous noiseless measurements are involved. We have chosen not to pursue this approach here, as a fixed edge in our approximate solution inherently implies zero curvature in the approximate solution~\cite{taylorSolutionClampedRectangular2004}.\par

\begin{figure}[!t]
	\centering
	\includegraphics[clip]{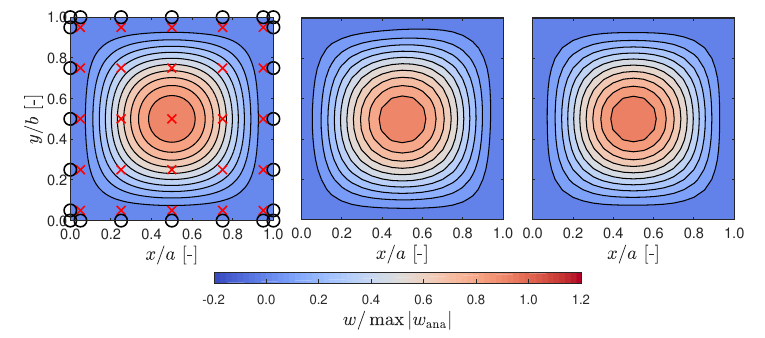} 
	\includegraphics[clip]{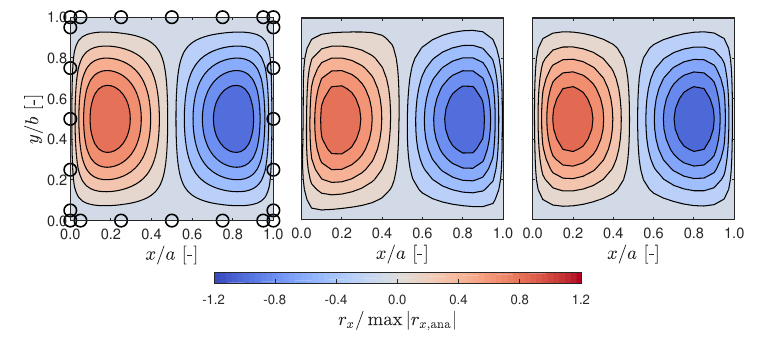} 
	\includegraphics[clip]{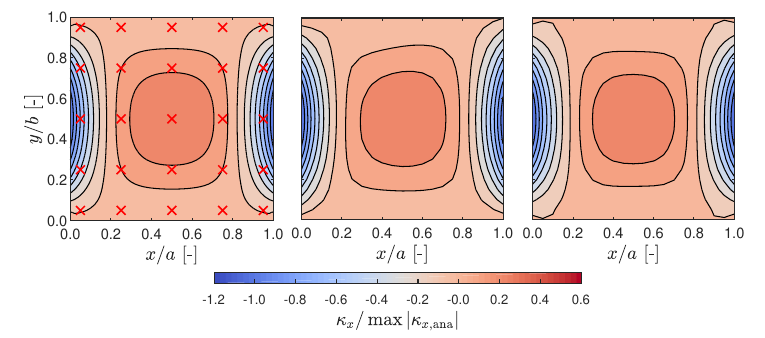} 
	\caption{Fixed plate with uniform loading: Normalized mean deflection $w$ (top), rotation $r_x$ (centre) and curvature $\kappa_x$  (bottom) - True values (left), prediction without boundary conditions (center) and with boundary conditions (right). The red crosses represent the training points and hollow circle the location of the imposed boundary conditions.}
	\label{fig:Ex2_Contour_Int}
\end{figure} 

\begin{figure}[!t]
	\centering
	\hspace*{-1.2cm}\includegraphics[clip,scale=1.0]{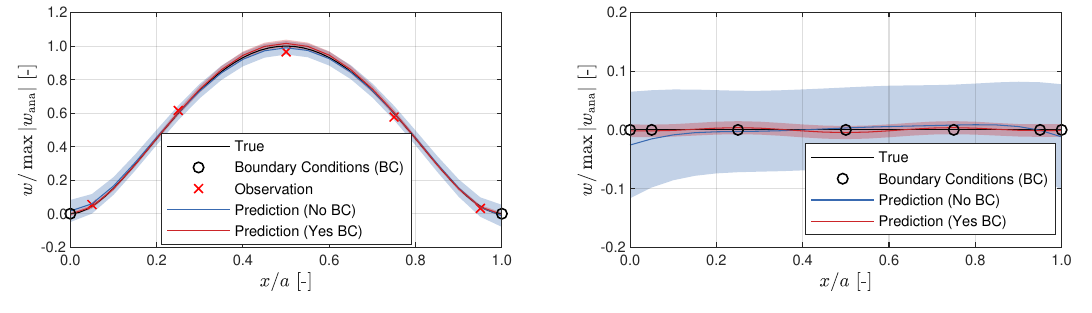}\vspace*{-0.3cm}
	\hspace*{-1.2cm}\includegraphics[clip,scale=1.0]{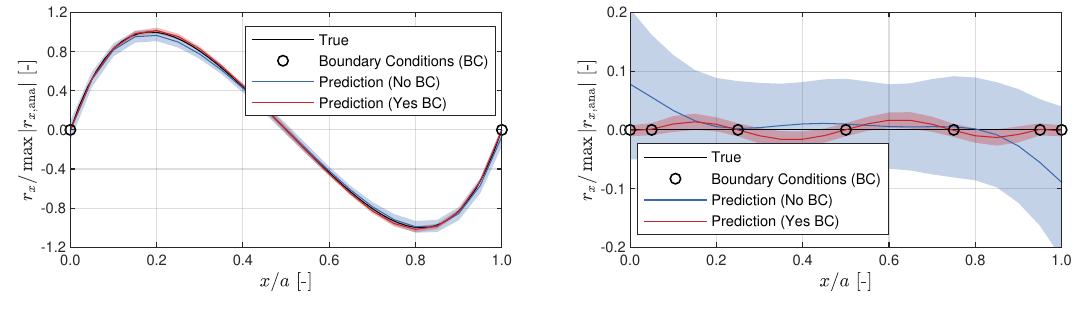}\vspace*{-0.3cm}
	\hspace*{-1.2cm}\includegraphics[clip,scale=1.0]{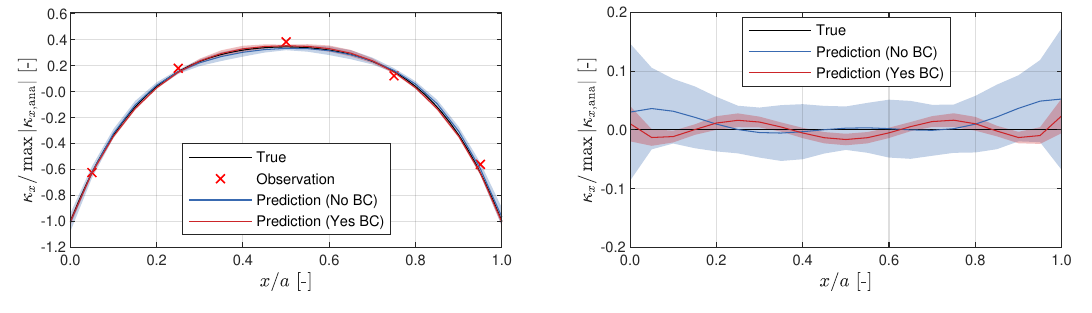}\vspace*{-0.3cm}
	%\hspace*{-1.2cm}\includegraphics[clip,scale=1.0]{Figures//Combined_Figure11//Ex2_Line_w_Rx_Kx.pdf}
	
	\caption{Fixed plate with uniform loading: deflection $w$ (top), rotation $r_x$ (centre) and curvature $\kappa_x$ (bottom) at the centreline $y=0.5b$ (left) and at the support $y=0b$ (see Fig.~\ref{fig:Ex2_Contour_Int}) for the cases with and without boundary conditions (BC). The true values are the analytical solution; the observations (red crosses) include the true values and additional measurement noise, while the boundary conditions (hollow circles) do not include measurement noise; the prediction is obtained using the GP model. The shaded areas in the corresponding colours are the 99\% confidence interval for the prediction.}
	\label{fig:Ex2_Line}
\end{figure} 

Figure~\ref{fig:Ex2_Histogram_D_MCMC} shows the learned rigidity using both MLE (point estimates $\hat{D}^e$) and MCMC (full probability $p(D)$) for the two cases with and without boundary conditions (BC). In both cases, the rigidity is predicted close to the true value. Additionally, Fig.~\ref{fig:Ex2_Parameters} depicts the MCMC histograms of the kernel hyperparameters and their corresponding pairwise scatters with correlation coefficients. The MCMC mean $\overline{D^e}$ for the case without boundary conditions results with less than 1\% difference from the true value, while the mean deviates by 1.9\% of the true value for the case with boundary conditions.  Thus, very good results are obtained in both cases. It may seem counter-intuitive that by providing boundary conditions, and thus, more data, a worse rigidity estimate is obtained. One possible reason is that the learning is driven by the likelihood terms corresponding to the noiseless data at boundary conditions, which constrains the algorithm. In addition, the length scale $l_x$ and $l_y$ for the case with boundary conditions are smaller, resulting in lower correlation between the boundary conditions and nearby observations. A solution to this may be to include the BCs directly into the kernel formulation, instead of providing it as artificial noiseless measurement points. The MLE estimates of the rigidity are relatively worse than MCMC means w.r.t. the true value. This is attributed to the effect of large jitter $\varepsilon$ required when using a gradient-based optimizer for the MLE.  \par
Next, we use MCMC for prediction. Figure~\ref{fig:Ex2_Contour_Int} depicts the mean prediction of the deflections $w$, rotation $r_x$ and curvature $k_x$. Further, Fig.~\ref{fig:Ex2_Line} depicts the mean and 99\% confidence interval of all these quantities at the centreline $y=0.5b$ (left) and at the boundary $y=0$ (right). Generally, very good correspondence is reached for both observed (deflections $w$ and curvature $k_x$) and unobserved $r_x$ quantities. The case with boundary conditions resulted in slightly better mean values. To this end, the zero deflections and rotation at the fixed edge $y=0$ (Fig.~\ref{fig:Ex2_Line}, top and center) are significantly better captured when imposed. Moreover, the case with boundary conditions resulted in less uncertainty. It is interesting to note that, despite the boundary conditions being applied as noiseless measurements for the displacements and rotations, there is still predictive uncertainty at these locations. This uncertainty arises from the jitter $\varepsilon$ added to the covariance matrix to ensure numerical stability when inverting it (see Sec.~\ref{sec:Learning}).\par
The results demonstrate that the model is capable of imposing artificial boundary conditions, improving its overall quality at the expense of less accurate response predictions towards the center of the plate. Although the deviation of the rigidity prediction for the case with boundary conditions is not so significant practically (less than 2\% for noisy data), a possible way to alleviate it could be by imposing the boundary conditions directly on the kernel~\cite{gulianGaussianProcessRegression2022}. This approach follows an inverse order, placing the GP model on the loads and deriving the other covariance kernels via Green's functions. However, this constrains the GP model to a specific plate model, as the integration of the covariance kernel requires boundary condition information to derive particular solutions.

\section{Conclusions}\label{sec:Conclusion}
We presented a physics-informed GP scheme for inference of structural rigidity and physical quantities of Kirchhoff-Love plates based on heterogeneous noisy measurements. A model was constructed by placing a GP prior on the deflections and propagating the prior through the linear partial differential equations of the plate mechanical model, yielding a multi-output GP model. Analytical cross-covariance functions were derived to construct the covariance matrix. The predictive distributions of the flexural rigidity and plate quantities were obtained by MCMC sampling and/or MLE estimates. \par 
The numerical experiments of a simply-supported plate and a fixed plate demonstrated the model's ability for inference of the rigidity probability distribution. It was shown that considering heterogeneous measurements (e.g. deflections and curvatures) within a physics-informed setting improves the learned rigidity in terms of both the mean and spread of the distribution. Nevertheless, it is important to note that, in practical terms, some quantities (e.g., deflections and rotations) are easier to measure than others (e.g., curvatures from optics~\cite{kaoFamilyGratingTechniques1982}) or can be obtained indirectly (e.g., indirect curvature from strains). Further, the model proved capable of predicting both observed and unobserved plate quantities (deformations and internal forces) at unobserved locations. Prescribing boundary conditions as noiseless measurements improved the prediction of the physical quantities, but not necessarily the learned rigidity.\par

Applications of the presented model are foreseen in SHM within the fields of structural and mechanical engineering, e.g. steel box girders of bridges. The current study is based on numerical experiments, assuming Gaussian noise, which may not reflect real-world conditions. Future studies could involve real experiments to validate the methodology. Additionally, there could be modeling errors since the Kirchhoff-Love theory might not accurately represent reality, particularly in the case of thick plates. This limitation arises because the multi-output GP is constrained by the partial differential equation. Modeling uncertainty could potentially be captured by placing additional GP priors on the individual physical terms~\cite{girolamiStatisticalFiniteElement2021}. Future research may involve appropriate kernel selection based on loading and geometry type, and incorporating boundary conditions directly at a kernel level. This is particularly relevant for fields where discontinuities are introduced, such as point force loading.. Alternatively, methods for automatic kernel construction can be applied~\cite{calandraManifoldGaussianProcesses2016}.
 Integration of the presented model within a system of other structural elements (e.g. beams) or within a reliability analysis remain a viable outlook.

\section*{Data Availability Statement}
Some or all data, models, or code generated or used during the study are available in a repository online in accordance with funder data retention policies: \href{https://github.com/IgorKavrakov/PlateGP}{github.com/IgorKavrakov/PlateGP}
\section*{Acknowledgments}\label{sec:Acknowledgments}
IK gratefully acknowledges the support by the German
Research Foundation (DFG) [Project No. 491258960], Darwin College and the Department of Engineering, University of Cambridge. 

\bibliographystyle{arxiv} 
\bibliography{references}

\newpage
\appendix
\section{Covariance functions}\label{App:CovFunc}
\begin{equation*}\label{eq:Kernels}
	\begin{aligned}
		k_{wr_x}=&\frac{\partial}{\partial x'}k_{ww}, \\ k_{wr_y}=&\frac{\partial}{\partial y'}k_{ww},\\
		k_{w\kappa_x}=&-\frac{\partial^2}{\partial x'^2}k_{ww}, \\
		k_{w\kappa_y}=&-\frac{\partial^2}{\partial y'^2}k_{ww},\\
		k_{w\kappa_{xy}}=&-\frac{2\partial^2}{\partial x'\partial y'}k_{ww}, \\
		k_{wq}^D=&D\left(\frac{\partial^{4}}{\partial x'^{4}}+\frac{2\partial^{4}}{\partial x'^{2}\partial y'^{2}}+\frac{\partial^{4}}{\partial y'^{4}}\right)k_{ww},\\
		k_{wQ_{x}}^D=&-D\left(\frac{\partial^3}{\partial x'^3}+\frac{\partial^3}{\partial x'\partial y'^2}\right)k_{ww},\\ 
		k_{wQ_{y}}^D=&-D\left(\frac{\partial^3}{\partial x'^2\partial y'}+\frac{\partial^3}{\partial y'^3}\right)k_{ww},\\
		k_{wM_{x}}^D=&-D\left(\frac{\partial^2}{\partial x'^2}+\nu\frac{\partial^2}{\partial y'^2}\right)k_{ww}, \\ 
		k_{wM_{y}}^D=&-D\left(\frac{\partial^2}{\partial y'^2}+\nu\frac{\partial^2}{\partial x'^2}\right)k_{ww},\\
		k_{wM_{xy}}^D=&D(1-\nu)\frac{\partial^2}{\partial x' \partial y'}k_{ww}, \\
		k_{r_xr_x}=&\frac{\partial^2}{\partial x \partial x^\prime}k_{ww},\\ k_{r_xr_y}=&\frac{\partial^2}{\partial x\partial y^\prime}k_{ww},\\
		k_{r_x\kappa_x}=&-\frac{\partial^3}{\partial x \partial {x^\prime}^2}k_{ww},\\ k_{r_x\kappa_y}=&-\frac{\partial^3}{\partial x\partial {y^\prime}^2}k_{ww},\\
		k_{r_x\kappa_{xy}}=&-\frac{2\partial^3}{\partial x \partial x^\prime \partial y^\prime}k_{ww}, \\ k_{r_xq}^D=&D\left(\frac{\partial^{5}}{\partial x\partial {x^\prime}^{4}}+\frac{2\partial^{5}}{\partial x\partial {x^\prime}^{2}\partial {y^\prime}^{2}}+\frac{\partial^{5}}{\partial x\partial {y^\prime}^{4}}\right)k_{ww},\\
		k_{r_xQ_{x}}^D=&-D\left(\frac{\partial^4}{\partial x\partial {x^\prime}^3}+\frac{\partial^4}{\partial x\partial {x^\prime}\partial {y^\prime}^2}\right)k_{ww},\\ k_{r_xQ_{y}}^D=&-D\left(\frac{\partial^4}{\partial x\partial {x^\prime}^2\partial y^\prime}+\frac{\partial^4}{\partial x \partial {y^\prime}^3}\right)k_{ww},\\
		k_{r_xM_{x}}^D=&-D\left(\frac{\partial^3}{\partial x\partial {x^\prime}^2}+\nu\frac{\partial^3}{\partial x\partial {y^\prime}^2}\right)k_{ww}, \\
		k_{r_xM_{y}}^D=&-D\left(\frac{\partial^3}{\partial x\partial {y^\prime}^2}+\nu\frac{\partial^3}{\partial x\partial {x^\prime}^2}\right)k_{ww},\\
		k_{r_xM_{xy}}^D=&D(1-\nu)\frac{\partial^3}{\partial x\partial x^\prime \partial y^\prime}k_{ww}, \\
		k_{r_yr_y}=&\frac{\partial^2}{\partial y\partial y^\prime}k_{ww},\\
	\end{aligned}
\end{equation*}
\begin{equation*}\label{eq:Kernels1}
	\begin{aligned}
		k_{r_y\kappa_x}=&-\frac{\partial^3}{\partial y\partial x'^2}k_{ww}, \\ k_{r_y\kappa_y}=&-\frac{\partial^3}{\partial y\partial {y^\prime}^2}k_{ww},\\
		k_{r_y\kappa_{xy}}=&-\frac{2\partial^3}{\partial y\partial x^\prime\partial y^\prime}k_{ww}, \\
		k_{r_yq}^D=&D\left(\frac{\partial^{5}}{\partial y\partial {x^\prime}^{4}}+\frac{2\partial^{5}}{\partial y\partial {x^\prime}^{2}\partial {y^\prime}^{2}}+\frac{\partial^{5}}{\partial y\partial {y^\prime}^{4}}\right)k_{ww},\\
		k_{r_yQ_{x}}^D=&-D\left(\frac{\partial^4}{\partial y\partial {x^\prime}^3}+\frac{\partial^4}{\partial y\partial {x^\prime}\partial {y^\prime}^2}\right)k_{ww},\\ k_{r_yQ_{y}}^D=&-D\left(\frac{\partial^4}{\partial y\partial {x^\prime}^2\partial y^\prime}+\frac{\partial^4}{\partial y \partial {y^\prime}^3}\right)k_{ww},\\
		k_{r_yM_{x}}^D=&-D\left(\frac{\partial^3}{\partial y\partial {x^\prime}^2}+\nu\frac{\partial^3}{\partial y\partial {y^\prime}^2}\right)k_{ww}, \\ k_{r_yM_{y}}^D=&-D\left(\frac{\partial^3}{\partial y\partial {y^\prime}^2}+\nu\frac{\partial^3}{\partial y\partial {x^\prime}^2}\right)k_{ww},\\
		k_{r_yM_{xy}}^D=&D(1-\nu)\frac{\partial^3}{\partial y\partial x^\prime \partial y^\prime}k_{ww}, \\
		k_{\kappa_x\kappa_x}=&\frac{\partial^4}{\partial x^2\partial {x^\prime}^2}k_{ww}, \\
		k_{\kappa_x\kappa_y}=&\frac{\partial^4}{\partial x^2\partial {y^\prime}^2}k_{ww},\\
		k_{\kappa_x\kappa_{xy}}=&\frac{2\partial^4}{\partial x^2\partial x^\prime\partial y^\prime}k_{ww},\\
		k_{\kappa_xq}^D=&-D\left(\frac{\partial^{6}}{\partial x^2\partial {x^\prime}^{4}}+\frac{2\partial^{6}}{\partial x^2\partial {x^\prime}^{2}\partial {y^\prime}^{2}}+\frac{\partial^{6}}{\partial x^2\partial {y^\prime}^{4}}\right)k_{ww},\\
		k_{\kappa_xQ_{x}}^D=&D\left(\frac{\partial^5}{\partial x^2\partial {x^\prime}^3}+\frac{\partial^5}{\partial x^2\partial {x^\prime}\partial {y^\prime}^2}\right)k_{ww},\\ k_{\kappa_xQ_{y}}^D=&D\left(\frac{\partial^5}{\partial x^2\partial {x^\prime}^2\partial y^\prime}+\frac{\partial^5}{\partial x^2 \partial {y^\prime}^3}\right)k_{ww},\\
		k_{\kappa_xM_{x}}^D=&D\left(\frac{\partial^4}{\partial x^2\partial {x^\prime}^2}+\nu\frac{\partial^4}{\partial x^2\partial {y^\prime}^2}\right)k_{ww}, \\
		k_{\kappa_xM_{y}}^D=&D\left(\frac{\partial^4}{\partial x^2\partial {y^\prime}^2}+\nu\frac{\partial^4}{\partial ^2x\partial {x^\prime}^2}\right)k_{ww},\\
		k_{\kappa_xM_{xy}}^D=&-D(1-\nu)\frac{\partial^4}{\partial x^2\partial x^\prime \partial y^\prime} k_{ww}, \\
		k_{\kappa_y\kappa_y}=&\frac{\partial^4}{\partial y^2\partial {y^\prime}^2}k_{ww},&\\
		k_{\kappa_y\kappa_{xy}}=&\frac{2\partial^4}{\partial y^2\partial x^\prime\partial y^\prime}k_{ww},\\
		k_{\kappa_yq}^D=&-D\left(\frac{\partial^{6}}{\partial y^{2}\partial {x^\prime}^4}+\frac{2\partial^{6}}{\partial y^2\partial {x^\prime}^{2}\partial {y^\prime}^{2}}+\frac{\partial^{6}}{\partial y^2\partial {y^\prime}^{4}}\right)k_{ww},\\
		k_{\kappa_yQ_{x}}^D=&D\left(\frac{\partial^5}{\partial y^2\partial {x^\prime}^3}+\frac{\partial^5}{\partial y^2\partial {x^\prime}\partial {y^\prime}^2}\right)k_{ww},\\ 
		k_{\kappa_yQ_{y}}^D=&D\left(\frac{\partial^5}{\partial y^2\partial {x^\prime}^2\partial y^\prime}+\frac{\partial^5}{\partial y^2 \partial {y^\prime}^3}\right)k_{ww},\\
	\end{aligned}
\end{equation*}
\begin{equation*}\label{eq:Kernels2}
	\begin{aligned}
		k_{\kappa_yM_{x}}^D=&D\left(\frac{\partial^4}{\partial y^2\partial {x^\prime}^2}+\nu\frac{\partial^4}{\partial y^2\partial {y^\prime}^2}\right)k_{ww}, \\
		k_{\kappa_yM_{y}}^D=&D\left(\frac{\partial^4}{\partial y^2\partial {y^\prime}^2}+\nu\frac{\partial^4}{\partial y^2\partial {x^\prime}^2}\right)k_{ww}\\
		k_{\kappa_yM_{xy}}^D=&-D(1-\nu)\frac{\partial^4}{\partial y^2\partial x^\prime \partial y^\prime}k_{ww}, \\
		k_{\kappa_{xy}\kappa_{xy}}=&\frac{4\partial^4}{\partial x\partial y\partial x^\prime\partial y^\prime}k_{ww},\\
		k_{\kappa_{xy}q}^D=&-D\left(\frac{2\partial^{6}}{\partial x \partial y\partial {x^\prime}^4}+\frac{4\partial^{6}}{\partial x\partial y\partial {x^\prime}^{2}\partial {y^\prime}^{2}}+\frac{2\partial^{6}}{\partial x \partial y\partial {y^\prime}^{4}}\right)k_{ww},\\
		k_{\kappa_{xy}Q_{x}}^D=&2D\left(\frac{\partial^5}{\partial x\partial y\partial {x^\prime}^3}+\frac{\partial^5}{\partial x\partial y\partial {x^\prime}\partial {y^\prime}^2}\right)k_{ww},\\ k_{\kappa_{xy}Q_{y}}^D=&2D\left(\frac{\partial^5}{\partial x\partial y\partial {x^\prime}^2\partial y^\prime}+\frac{\partial^5}{\partial x\partial y \partial {y^\prime}^3}\right)k_{ww},\\
		k_{\kappa_{xy}M_{x}}^D=&2D\left(\frac{\partial^4}{\partial x\partial y\partial {x^\prime}^2}+\nu\frac{\partial^4}{\partial x\partial y\partial {y^\prime}^2}\right)k_{ww}, \\ k_{\kappa_{xy}M_{y}}^D=&2D\left(\frac{\partial^4}{\partial x\partial y\partial {y^\prime}^2}+\nu\frac{\partial^4}{\partial x\partial y\partial {x^\prime}^2}\right)k_{ww},\\
		k_{\kappa_{xy}M_{xy}}^D=&-2D(1-\nu)\frac{\partial^4}{\partial x\partial y\partial x^\prime \partial y^\prime}k_{ww}, \\
		k_{qq}^D=&D^2\Bigg(\frac{\partial^{8}}{\partial  x^4 \partial{x^\prime}^{4}}+\frac{2\partial^{8}}{\partial  x^4\partial {x^\prime}^{2}\partial {y^\prime}^{2}}+\frac{\partial^{8}}{\partial  x^4\partial {y^\prime}^{4}}+\frac{2\partial^{8}}{\partial  x^2 \partial  y^2 \partial {x^\prime}^{4}}\\
		&+\frac{4\partial^{8}}{\partial  x^2 \partial  y^2\partial {x^\prime}^{2}\partial {y^\prime}^{2}}+\frac{2\partial^{8}}{\partial  x^2 \partial  y^2\partial {y^\prime}^{4}}+\frac{\partial^{8}}{\partial  y^4 \partial{x^\prime}^{4}}\\
		&+\frac{2\partial^{8}}{\partial  y^4\partial {x^\prime}^{2}\partial {y^\prime}^{2}}+\frac{\partial^{8}}{\partial  y^4\partial {y^\prime}^{4}}\Bigg)k_{ww},\\
		k_{qQ_{x}}^D=&-D^2\Bigg(\frac{\partial^7}{\partial x^4\partial {x^\prime}^3}+\frac{\partial^7}{\partial x^4 \partial x^\prime\partial {y^\prime}^2}+\frac{2\partial^7}{\partial x^2\partial y^2\partial {x^\prime}^3}\\
		&+\frac{2\partial^7}{\partial x^2 \partial y^2 \partial x^\prime\partial {y^\prime}^2}+\frac{\partial^7}{\partial y^4\partial {x^\prime}^3}+\frac{\partial^7}{\partial y^4 \partial x^\prime\partial {y^\prime}^2}\Bigg)k_{ww},\\ 
		k_{qQ_{y}}^D=&-D^2\Bigg(\frac{\partial^7}{\partial x^4\partial {x^\prime}^2\partial {y^\prime}}+\frac{\partial^7}{\partial x^4\partial {y^\prime}^3}+\frac{2\partial^7}{\partial x^2\partial y^2\partial {x^\prime}^2\partial {y^\prime}}\\
		&+\frac{2\partial^7}{\partial x^2 \partial y^2 \partial {y^\prime}^3}+\frac{\partial^7}{\partial y^4\partial {x^\prime}^2\partial {y^\prime}}+\frac{\partial^7}{\partial y^4 \partial {y^\prime}^3}\Bigg)k_{ww},\\
		k_{qM_{x}}^D=&-D^2\Bigg(\frac{\partial^6}{\partial x^4\partial {x^\prime}^2}+\nu\frac{\partial^6}{\partial x^4 \partial {y^\prime}^2}+\frac{2\partial^6}{\partial x^2\partial y^2\partial {x^\prime}^2}\\
		&+\nu\frac{2\partial^6}{\partial x^2 \partial y^2 \partial {y^\prime}^2}+\frac{\partial^6}{\partial y^4\partial {x^\prime}^2}+\nu\frac{\partial^6}{\partial y^4\partial {y^\prime}^2}\Bigg)k_{ww},\\ 
		k_{qM_{y}}^D=&-D^2\Bigg(\frac{\partial^6}{\partial x^4\partial {y^\prime}^2}+\nu\frac{\partial^6}{\partial x^4 \partial {x^\prime}^2}+\frac{2\partial^6}{\partial x^2\partial y^2\partial {y^\prime}^2}\\
		&+\nu\frac{2\partial^6}{\partial x^2 \partial y^2 \partial {x^\prime}^2}+\frac{\partial^6}{\partial y^4\partial {y^\prime}^2}+\nu\frac{\partial^6}{\partial y^4\partial {x^\prime}^2}\Bigg)k_{ww},\\
	\end{aligned}
\end{equation*}
\begin{equation*}\label{eq:Kernels4}
	\begin{aligned}
		k_{qM_{xy}}^D=&D^2(1-\nu)\Bigg(\frac{\partial^6}{\partial x^4\partial x^\prime\partial y^\prime}+\frac{2\partial^6}{\partial x^2\partial y^2\partial x^\prime\partial y^\prime}+\frac{\partial^6}{\partial y^4 \partial x^\prime\partial y^\prime}\Bigg)k_{ww}, \\
		k_{Q_{x}Q_{x}}^D=&D^2\Bigg(\frac{\partial^6}{\partial x^3\partial {x^\prime}^3}+\frac{\partial^6}{\partial x^3 \partial {x^\prime}\partial {y^\prime}^2}+\frac{\partial^6}{\partial x\partial y^2\partial {x^\prime}^3}+\frac{\partial^6}{\partial x\partial y^2 \partial {x^\prime}\partial {y^\prime}^2}\Bigg)k_{ww},\\ 
		k_{Q_{x}Q_{y}}^D=&D^2\Bigg(\frac{\partial^6}{\partial x^3\partial {x^\prime}^2\partial y^\prime}+\frac{\partial^6}{\partial x^3 \partial {y^\prime}^3}+\frac{\partial^6}{\partial x\partial y^2\partial {x^\prime}^2\partial y^\prime}+\frac{\partial^6}{\partial x\partial y^2 \partial {y^\prime}^3}\Bigg)k_{ww},\\
		k_{Q_{x}M_{x}}^D=&D^2\Bigg(\frac{\partial^5}{\partial x^3\partial {x^\prime}^2}+\nu\frac{\partial^5}{\partial x^3\partial {y^\prime}^2}+\frac{\partial^5}{\partial x\partial y^2{x^\prime}^2}+\nu\frac{\partial^5}{\partial x\partial y^2\partial {y^\prime}^2}\Bigg)k_{ww}, \\
		k_{Q_{x}M_{y}}^D=&D^2\Bigg(\frac{\partial^5}{\partial x^3\partial {y^\prime}^2}+\nu\frac{\partial^5}{\partial x^3\partial {x^\prime}^2}+\frac{\partial^5}{\partial x\partial y^2{y^\prime}^2}+\nu\frac{\partial^5}{\partial x\partial y^2\partial {x^\prime}^2}\Bigg)k_{ww},\\
		k_{Q_{x}M_{xy}}^D=&-D^2(1-\nu)\left(\frac{\partial^5}{\partial x^3 \partial x^\prime \partial y^\prime}+\frac{\partial^5}{\partial x \partial y^2\partial x^\prime \partial y^\prime}\right)k_{ww}, \\
		k_{Q_{y}Q_{y}}^D=&D^2\Bigg(\frac{\partial^6}{\partial x^2\partial y \partial {x^\prime}^2\partial {y^\prime}}+\frac{\partial^6}{\partial x^2\partial y\partial {y^\prime}^3}+\frac{\partial^6}{\partial y^3 \partial {x^\prime}^2\partial {y^\prime}}+\frac{\partial^6}{\partial y^3\partial {y^\prime}^3}\Bigg)k_{ww},\\ 
		k_{Q_{y}M_{x}}^D=&D^2\Bigg(\frac{\partial^5}{\partial x^2\partial y\partial{x^\prime}^2}+\nu\frac{\partial^5}{\partial x^2\partial y\partial {y^\prime}^2}+\frac{\partial^5}{\partial y^3\partial {x^\prime}^2}+\nu\frac{\partial^5}{\partial y^3\partial {y^\prime}^2}\Bigg)k_{ww},\\
		k_{Q_{y}M_{y}}^D=&D^2\Bigg(\frac{\partial^5}{\partial x^2\partial y\partial{y^\prime}^2}+\nu\frac{\partial^5}{\partial x^2\partial y\partial {x^\prime}^2}+\frac{\partial^5}{\partial y^3\partial {y^\prime}^2}+\nu\frac{\partial^5}{\partial y^3\partial {x^\prime}^2}\Bigg)k_{ww},\\
		k_{Q_{y}M_{xy}}^D=&-D^2(1-\nu)\left(\frac{\partial^5}{\partial x^2 \partial y\partial x^\prime \partial y^\prime}+\frac{\partial^5}{\partial y^3 \partial x^\prime \partial y^\prime}\right)k_{ww},\\
		k_{M_{x}M_{x}}^D=&D^2\left(\frac{\partial^4}{\partial x^2\partial {x^\prime}^2}+\nu\frac{\partial^4}{\partial x^2\partial {y^\prime}^2}+\nu\frac{\partial^4}{\partial y^2\partial {x^\prime}^2}+\nu^2\frac{\partial^4}{\partial y^2\partial {y^\prime}^2}\right)k_{ww}, \\
		k_{M_{x}M_{y}}^D=&D^2\left(\frac{\partial^4}{\partial x^2\partial {y^\prime}^2}+\nu\frac{\partial^4}{\partial x^2\partial {x^\prime}^2}+\nu\frac{\partial^4}{\partial y^2\partial {y^\prime}^2}+\nu^2\frac{\partial^4}{\partial y^2\partial {x^\prime}^2}\right)k_{ww}\\
		k_{M_{x}M_{xy}}^D=&-D^2(1-\nu)\left(\frac{\partial^4}{\partial x^2 \partial x^\prime \partial y^\prime}+\nu\frac{\partial^4}{\partial y^2 \partial x^\prime \partial y^\prime}\right)k_{ww},\\\
		k_{M_{y}M_{y}}^D=&D^2\left(\frac{\partial^4}{\partial y^2\partial {y^\prime}^2}+\nu\frac{\partial^4}{\partial y^2\partial {x^\prime}^2}+\nu\frac{\partial^4}{\partial x^2\partial {y^\prime}^2}+\nu^2\frac{\partial^4}{\partial x^2\partial {x^\prime}^2}\right)k_{ww}, \\
		k_{M_{y}M_{xy}}^D=&-D^2(1-\nu)\left(\frac{\partial^4}{\partial y^2 \partial x^\prime \partial y^\prime}+\nu\frac{\partial^4}{\partial x^2 \partial x^\prime \partial y^\prime}\right)k_{ww},\\\
		k_{M_{xy}M_{xy}}^D=&D^2(1-\nu)^2\frac{\partial^4}{\partial x \partial y \partial x^\prime \partial y^\prime}k_{ww}.\\
	\end{aligned}
\end{equation*}

\section{Metropolis-Hastings algorithm}\label{App:MH}
%
%{\color{red} TODO:
%
%\begin{itemize}
%    \item maybe add explanation?
%    \item this requires the packages "mathtools, algorithm, algorithmicx, algpseudocode". Not sure if that is ok, or if I should remove \& adapt to basic ones
%\end{itemize}
%}

\begin{algorithm}[H]
	\algnewcommand{\LineComment}[1]{\State // #1}
	\caption{Markov chain Monte Carlo using the Metropolis Hastings algorithm}
	\textbf{Inputs:}
	\begin{algorithmic}[0]
	    \LineComment{Initial hyperparameters, parameter posterior model, a parameter proposal distribution $g$ and the MCMC sample length $N_s$ and burn-in size $N_b$}
		\State{$\boldsymbol{\theta}_0^e = \left( A, l_x, l_y, D, \boldsymbol{\sigma}^2 \right)$, $\boldsymbol{\theta}^e_0\in\mathbb{R}^{1\times N_p}$}
		\State{$p(\boldsymbol{\theta}^e|\boldsymbol{z}, \boldsymbol{X}) \propto p(\boldsymbol{z}|\boldsymbol{X},\boldsymbol{\theta}^e) p(\boldsymbol{\theta}^e)$}
		\State{$g(\boldsymbol{\theta}) = \mathcal{N}\left( \mu_{\boldsymbol{\theta}}, \Sigma_{\boldsymbol{\theta}} \right)$}
		\State{$N_s$, $N_b$}
	\end{algorithmic}
	\textbf{Procedure:}
	\begin{algorithmic}[0]
		\For{$i=0,1,2,..., \ (N_s+N_b-1)$}
		\LineComment{Sample new candidate parameters based on parameter proposal distribution}
		\State{$\boldsymbol{\theta}^e_* \sim g(\boldsymbol{\theta}^e_i)$}
		\LineComment{Sample random number from uniform distribution}
		\State{$a \sim \mathcal{U}(0,1)$} 
		\LineComment{Calculate the Metropolis Hastings acceptance ratio}
		\State{$r = \text{min} \Bigg\{  1, \, \dfrac{p(\boldsymbol{\theta}^e_*|\boldsymbol{z},\boldsymbol{X}) g(\boldsymbol{\theta}_i^e|\boldsymbol{\theta}^e_*)}{p(\boldsymbol{\theta}_i^e|\boldsymbol{z},\boldsymbol{X}) g(\boldsymbol{\theta}^e_*|\boldsymbol{\theta}_i^e)} \Bigg\}$} 
		\LineComment{Update chain with new parameters}
		\State{$\boldsymbol{\theta}^e_{i+1} = $ $\begin{cases}
				\boldsymbol{\theta}^e_*, & \text{if  } (r \geqslant  a)\\
				\boldsymbol{\theta}_i^e, & \text{otherwise}
			\end{cases}$}
		\EndFor
		\LineComment{Discard initial $N_b$ points at low density locations}
		\State{$\boldsymbol{\theta}^e \xleftarrow{\mathrm{burn-in}} \boldsymbol{\theta}^e_{i=N_b, N_b+1,\ ...,\ (N_b+N_s-1)}$}
		
	\end{algorithmic}
	\textbf{Output:}
	\begin{algorithmic}[0]
	    \LineComment{The samples approximating the parameter posterior distribution}
		\State$p(\boldsymbol{\theta}^e|\boldsymbol{z}, \boldsymbol{X}) \approx \boldsymbol{\theta}^e\in\mathbb{R}^{N_s\times N_p}$ 
	\end{algorithmic}
	\label{Alg01}
\end{algorithm}

\end{document}